\documentclass[conference]{IEEEtran}
\IEEEoverridecommandlockouts
\usepackage{cite}
\usepackage{amsmath,amssymb,amsfonts}
\usepackage{graphicx}
\usepackage{textcomp}
\usepackage{multirow}
\usepackage{booktabs}
\usepackage{caption}
\usepackage{xcolor}
\usepackage{booktabs}
\usepackage{makecell}  
\usepackage{flushend}
\usepackage{tabularx}
\usepackage{footnote}
\usepackage{subfigure}
\usepackage{amsfonts}
\usepackage{amsmath}
\usepackage{amsthm}
\usepackage{algorithm}
\usepackage{algorithmic}
\usepackage{float}
\usepackage{bm}
\usepackage{xspace}
\usepackage{threeparttable}
\usepackage{amssymb}
\usepackage{url}
\usepackage[misc]{ifsym}
\usepackage{enumerate}
\usepackage{balance}
\usepackage{ulem}
\usepackage{subcaption}

\usepackage{mathrsfs}
\usepackage{caption}
\usepackage{graphicx}
\usepackage{float} 
\usepackage{subcaption}

\def\BibTeX{{\rm B\kern-.05em{\sc i\kern-.025em b}\kern-.08em
    T\kern-.1667em\lower.7ex\hbox{E}\kern-.125emX}}

\usepackage[
    colorlinks=true,
    linkcolor=red,      
    urlcolor=blue,      
    citecolor=green,    
    linkbordercolor=red 
]{hyperref}

\newtheorem{Def}{\textbf{Definition}}

\newtheorem*{Pro*}{Problem}

\newcommand{\model}{TrajDiff\xspace}

\begin{document}


\title{\model: Diffusion Bridge Network with Semantic Alignment for Trajectory Similarity Computation}

\author{
\IEEEauthorblockN{
Xiao Zhang{\textsuperscript{1}}, Xingyu Zhao{\textsuperscript{1}}, Hong Xia{\textsuperscript{1}}, Yuan Cao{\textsuperscript{1}}, Guiyuan Jiang{\textsuperscript{1}}, Junyu Dong{\textsuperscript{1}, Yanwei Yu{\textsuperscript{1}}\textsuperscript{\Letter}}}\\
\IEEEauthorblockA{
\textsuperscript{1}Faculty of Information Science and Engineering, Ocean University of China, Qingdao, China\\
\{zhangxiao4549, zhaoxingyu, xhong\}@stu.ouc.edu.cn, \\ \{cy8661, jiangguiyuan, dongjunyu,yuyanwei \}@ouc.edu.cn}\\
}

\maketitle

\begin{abstract}


With the proliferation of location-tracking technologies, massive volumes of trajectory data are continuously being collected. As a fundamental task in trajectory data mining, trajectory similarity computation plays a critical role in a wide range of real-world applications. However, existing learning-based methods face three challenges: First, they ignore the semantic gap between GPS and grid features in trajectories, making it difficult to obtain meaningful trajectory embeddings. Second, the noise inherent in the trajectories, as well as the noise introduced during grid discretization, obscures the true motion patterns of the trajectories. Third, existing methods focus solely on point-wise and pair-wise losses, without utilizing the global ranking information obtained by sorting all trajectories according to their similarity to a given trajectory. To address the aforementioned challenges, we propose a novel trajectory similarity computation framework, named \model. Specifically, the semantic alignment module relies on cross-attention and an attention score mask mechanism with adaptive fusion, effectively eliminating semantic discrepancies between data at two scales and generating a unified representation. Additionally, the DDBM-based Noise-robust Pre-Training introduces the transfer patterns between any two trajectories into the model training process, enhancing the model's noise robustness. Finally, the overall ranking-aware regularization shifts the model’s focus from a local to a global perspective, enabling it to capture the holistic ordering information among trajectories. Extensive experiments on three publicly available datasets show that \model consistently outperforms state-of-the-art baselines. In particular, it achieves an average HR@1 gain of 33.38\% across all three evaluation metrics and datasets. The code for our model is available at \textcolor{blue}{\url{https://github.com/zhx66741/TrajDiff}}.

\end{abstract}


\section{Introduction}
With the rapid development of positioning and sensing technologies, vast amounts of trajectory data are continuously collected from diverse sources~\cite{cluster1,cluster2,traj_cluster,survey}. These spatio-temporal trajectories are essential in various applications, including intelligent transportation systems~\cite{intelligent-transportation1,ITS}, urban planning~\cite{urban-planning}, and influenza monitoring~\cite{liugan}. Among the key tasks in trajectory data mining, trajectory similarity computation is a fundamental operation~\cite{yao2017trajectory,yao2018learning,survey1}, forming the basis for a wide range of downstream tasks such as trajectory clustering, retrieval, and anomaly detection. Therefore, accurate and efficient similarity computation is crucial for informed decision-making and driving advancements in these fields~\cite{tmn,yichang1,yichang2}.

Generally, trajectory similarity computation methods can be categorized into two classes: \textbf{\textit{heuristic measures}} and \textbf{\textit{learnable measures}}~\cite{trajcl}. Heuristic measures are typically based on rules defined manually to identify the optimal point match between two trajectories~\cite{grlstm}. For instance, the Symmetric Segment Path Distance (SSPD) is computed as the symmetric average of the minimum distances between corresponding segments of two trajectories. However, heuristic measures typically rely on dynamic programming or geometric computations, resulting in a time complexity that often reaches \( \mathcal{O}(n^2) \) or even higher ~\cite{distcret_frechet,survey2,survey3}, where $n$ represents the number of points in the trajectory. When processing large-scale trajectory datasets, this computational complexity creates a significant performance bottleneck. Furthermore, for a dataset containing $N$ trajectories, the number of required pairwise computations increases quadratically, scaling as $\mathcal{O}(N^2)$.
For example, computing pairwise Fr\'{e}chet distances for 1,000, 2,000, and 7,000 trajectories takes approximately 1.4 hours, 3.2 hours, and more than 13 hours, respectively.

\begin{figure}
    \begin{center}
    \includegraphics[width=\linewidth]{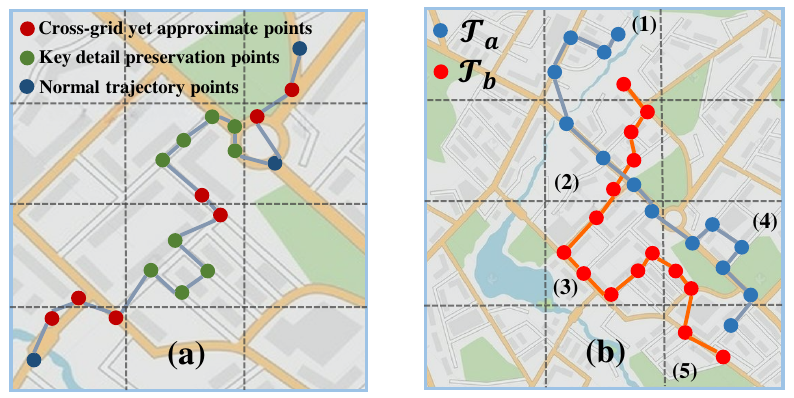}
    \caption{Many-to-one mapping issues caused by feature misalignment and noise interference. (a) All green points are projected onto the same grid, ignoring fine-grained information; the red points are geographically adjacent, but after discretization, they fall into different grids. (b) Two trajectories are projected onto the same grid sequence, yet exhibit differences in their activity regions and movement patterns.}
    \label{fig:challenge}
    \vspace{-4mm}
    \end{center}
\end{figure}

In contrast, learnable measures employ neural networks to map trajectories into $d$-dimensional space, transforming pairwise similarity computations into highly efficient vector operations, typically with $\mathcal{O}\left(1\right)$ time complexity. As an illustration, TrajCL~\cite{trajcl} adopts an attention-based encoder to model trajectories, achieving speed improvements of 21× and 25× compared to Hausdorff and Fr\'{e}chet distance computations, respectively. Motivated by this, we also propose a learning-based approach to enable fast and effective trajectory similarity computation.

Recent studies like KGTS~\cite{chen2024kgts}, TrajCL~\cite{trajcl}, and TrajGAT~\cite{trajgat} have advanced trajectory similarity computation using novel approaches. TrajGAT introduces a graph attention network framework that captures the structural dependencies among trajectory points by integrating both geographical semantics and topological connections, enabling more accurate trajectory similarity modeling. Building upon representation learning, TrajCL employs an unsupervised contrastive learning paradigm that generates augmented views of raw trajectories and optimizes a similarity-preserving objective to learn robust trajectory embeddings. Extending the modeling capacity with external knowledge, KGTS leverages knowledge graph embeddings and prompt-based learning, and jointly encodes trajectories and spatial grids through contrastive objectives to enhance trajectory similarity computation.

However, there are still three challenges in trajectory similarity computation. \textbf{\textit{First, the primary challenge arises from the misalignment between the fine-grained GPS features and coarse-grained grid features of trajectories.}} These alignment errors result in many-to-one mapping issues. On one hand, multiple trajectory points may be projected onto the same grid and assigned identical values, thereby overlooking fine-grained movement features such as vehicle acceleration and intersection turns. On the other hand, different trajectories might be mapped to the same grid sequence, appearing similar at the grid level but exhibiting significant differences in their finer details. As illustrated in Figure~\ref{fig:challenge}(a), the green point sequences capture fine-grained movement patterns. However, at the grid level, they are confined to the same grid cell, causing these detailed behaviors to be lost. Meanwhile, Fig.~\ref{fig:challenge}(b) illustrates that trajectory $\mathscr{T}_a$ is primarily active in regions (1) and (4), whereas trajectory $\mathscr{T}_b$ predominantly operates in regions (3) and (5). Despite both trajectories being mapped to the same grid sequence, they exhibit notable differences in their active regions and movement patterns within those regions.

\textbf{\textit{Second, the presence of noise in the data masks the intrinsic trajectory patterns, hindering the model's ability to learn precise and reliable representations.}} Trajectory data is inherently noisy, arising from factors such as the limitations of data collection devices, environmental interference, and variations in user behavior. Moreover, additional noise emerges during the transformation of continuous GPS features into discrete grid features. In such cases, when a trajectory point lies near the boundary of a grid cell, even a minor physical displacement or device error can be significantly amplified by the grid quantization mechanism. As depicted in Figure~\ref{fig:challenge}(a), the red point sequences exhibit a high degree of clustering within the fine-grained GPS coordinate space. However, following grid discretization, the spatial distance between their corresponding grid-based representations is substantially magnified.

\textbf{\textit{Third, existing methods often neglect the global ranking structure of trajectories, which is essential for capturing holistic similarities.}} Trajectory similarity computation and retrieval tasks fundamentally share the same core principles. The retrieval task involves identifying the most similar or top-$k$ samples from a dataset based on a given query item, which is conceptually aligned with a similar trajectory query, as both aim to achieve the same objective. However, mainstream approaches that use point-wise losses (e.g., MSE) or pair-wise losses (e.g., triplet loss) have significant limitations: the former treats individual samples in isolation, while the latter focuses solely on pairwise relationships, both of which fail to capture the overall ranking information across the dataset. This limitation is especially critical in practical applications. For instance, in the trajectory similarity query task, after applying the proposed overall ranking-aware regularization, we achieved average improvements of 15.47\%, 23.89\%, and 52.85\% in SSPD distance under HR@1 metric across three datasets, compared to using MSE loss alone.

To address the aforementioned challenges, we propose a novel trajectory similarity computation framework, named \textbf{\model}. Specifically, we propose the semantic alignment module to effectively bridge the semantic gap between features at two distinct scales, integrating them into a unified and expressive embedding. To further enhance its ability to model latent semantic transitions, we pre-train this module using Denoising Diffusion Bridge Models (DDBM), by learning to progressively denoise intermediate states conditioned on both endpoints, the module captures the underlying dynamics while preserving semantic consistency and improving robustness to input noise. Finally, we design the overall ranking-aware regularization to incorporate global ranking information into the model training, enabling it to capture the overall ranking patterns effectively.  Experiments on three public datasets show the superiority of our model. We achieve the maximum performance gain of 92.99\% on the T-Drive dataset under the SSPD evaluation metric.

We summarize the key contributions of this work as follows:
\begin{itemize}
    \item We propose a novel semantic alignment module that explicitly models and integrates information from two different scales, effectively eliminating the semantic gap. 
    
    \item We propose a novel pretraining method based on DDBM, which leverages both the forward and backward diffusion processes to learn robust and semantically consistent representations.
    

    \item We propose a novel overall ranking-aware regularization to optimize the embedding space from a global ranking perspective, improving query performance.

    \item We conduct extensive experiments on three real-world datasets to demonstrate the superiority of our \model over state-of-the-art baselines, achieving an average 53.58\% improvement under SSPD metrics across three datasets on HR@1. 
\end{itemize}

\section{Related work}
In this section, we review important related work on three critical aspects that significantly influence the problem addressed in this paper: grid embedding pretraining, trajectory feature aggregation, and self-supervised trajectory similarity computation. 

\subsection{Grid Embedding Pretraining}
In trajectory similarity computation, the geographical area is typically partitioned into non-overlapping grids of fixed size to obtain a coarse-grained representation of the overall trajectory. However, using only grid IDs as features is insufficient, prompting the adoption of various pretraining methods to learn high-dimensional grid representations. TrajCL ~\cite{trajcl} constructs the grids as a graph structure and pre-trains the grid embedding using the Node2Vec~\cite{node2vec}. TrajGAT also employs Node2Vec to obtain grid embeddings. Unlike TrajCL, it partitions the entire geographical area into a PR quadtree ~\cite{prtree} based on the spatial distribution of trajectory points. KGTS~\cite{chen2024kgts} pre-trains spatially-aware grid embeddings using a modified RotatE~\cite{sun2019rotate} based on complex-space rotation mechanisms. GRLSTM~\cite{zhou2023grlstm} pre-trains grid embedding using the TransH~\cite{wang2014knowledge} model on a knowledge graph that integrates trajectory and road network information. 

However, existing pretraining approaches face two major limitations: first, the pretrained grid embeddings remain fixed and are not updated during model training; second, these methods are generally graph-based, where the grid embeddings are updated by considering the connectivity of neighboring nodes, but they fail to account for skipping connections in real trajectories. To overcome these limitations, we abandon the pretraining paradigm and instead employ a projection layer that maps grid coordinates directly into a high-dimensional embedding space.

\subsection{GPS and Grid Feature Aggregation}
Effective fusion of GPS features and grid features is crucial for trajectory similarity computation. T3S~\cite{T3S} processes the GPS features and the grid features using LSTM \cite{lstm,lstm1} and a self-attention-based network. The final trajectory embedding is obtained by summing these two components. TrajCL leverages a dual-feature self-attention mechanism to fuse structural and spatial features of trajectories by separately computing and adaptively integrating their attention weights, enabling the learning of more comprehensive trajectory representations. In addition, some methods utilize only one type of feature. Traj2SimVec~\cite{zhang2020trajectory} introduces an efficient and robust model through innovative designs, including trajectory simplification, efficient indexing, sub-trajectory distance loss, and point matching loss. TMN~\cite{tmn} solely relies on the GPS features, and it proposes an attention-based matching mechanism for point-wise matching between different trajectories. The matching mechanism not only takes into account the sequential information of individual trajectories but also incorporates the interaction information between different trajectories for similarity, thereby improving the accuracy of similarity calculation. To learn more realistic grid embeddings, HHL-Traj~\cite{HHL-Traj} employs a hypergraph~\cite{hypergraph1} to model the grid space and a GRU with residual connections to obtain more discriminative trajectory embeddings.  Additionally, a deep hash network is used to transform the trajectory into binary hash codes, enabling fast similarity computation. 

\subsection{Self-supervised Trajectory Similarity Computation}
In recent years, trajectory similarity computation has increasingly embraced the self-supervised learning paradigm~\cite{byol,ag}, which enables the acquisition of more discriminative trajectory representations without relying on manual annotations. Existing self-supervised methods generally follow two main approaches: reconstruction-based and contrastive-based learning. t2vec~\cite{t2vec} employs a seq2seq~\cite{seq2seq, seq2seq1} architecture to encode low sampling rate trajectories into continuous vectors in an unsupervised setting, and achieves similarity modeling by reconstructing high sampling rate trajectories. In contrast, contrastive learning methods directly optimize representation similarity. CL-TSim~\cite{cl-tsim} adopts the SimCLR framework~\cite{simclr} for pre-training via contrastive learning, where positive and negative sample pairs are generated through trajectory downsampling and perturbation. Similarly, TrajCL is built upon the MoCo framework~\cite{moco}, enhancing the stability of contrastive learning through a momentum encoder and a negative sample queue, and then fine-tunes the model to obtain robust and discriminative trajectory representations.  Furthermore, KGST~\cite{chen2024kgts} proposes grid-aware data augmentation strategies and introduces Prompt Trajectory Embedding, which guides the encoder to capture structured semantic patterns in grid-based trajectories better.



\begin{figure*}[htbp]
    \centering
    \includegraphics[width=\textwidth]{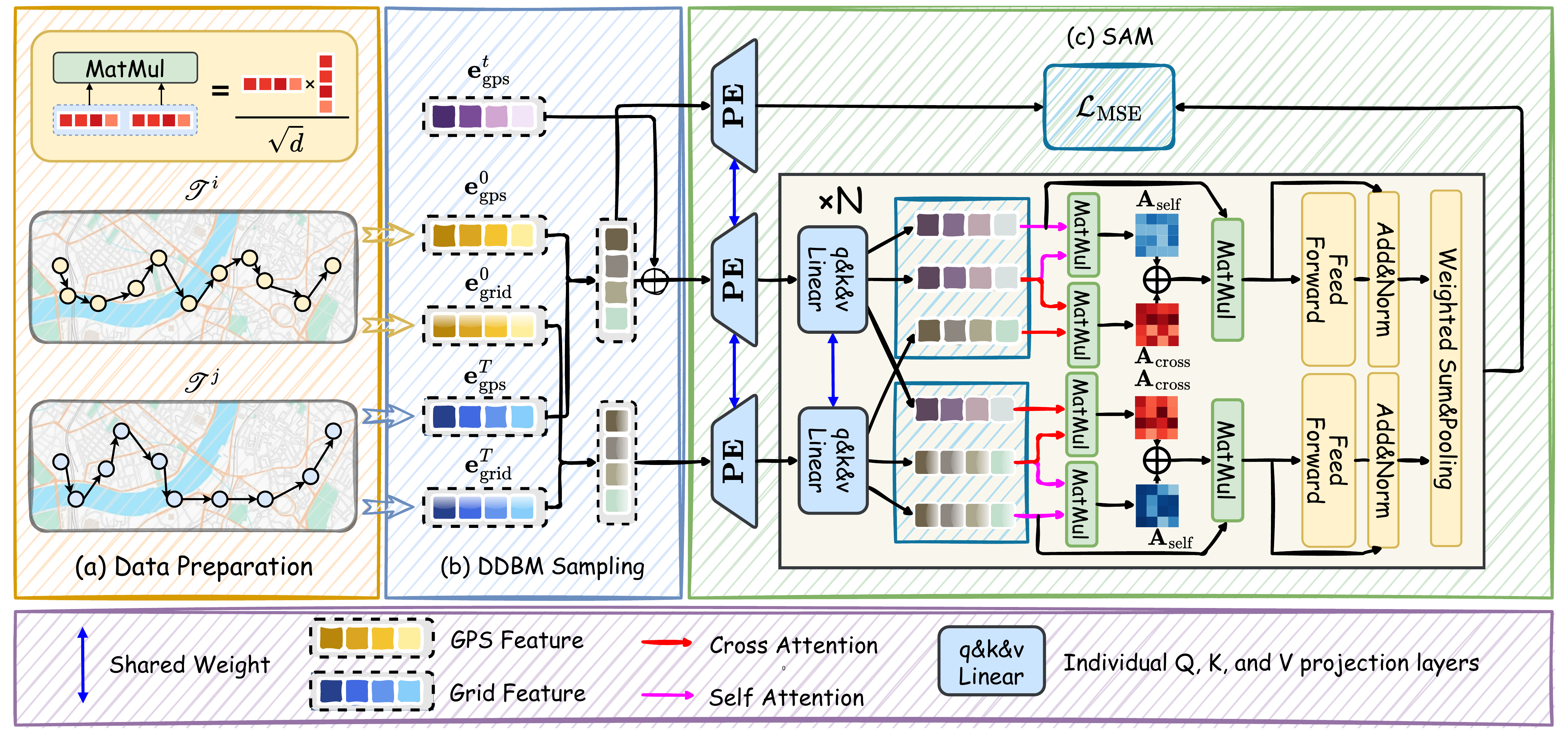}
    \caption{DDBM-based Noise-robust Pre-Training.}
    \label{fig:pretrain}
\end{figure*}

\section{PRELIMINARIES}
In this section, we introduce the foundational concepts and formally define the problem of trajectory similarity computation.
\begin{Def}[Trajectory]
    A trajectory is defined as an ordered sequence of spatio-temporal points: $\mathscr{T} = [p_1, p_2, \dots, p_n]$, where $p_i$ denotes the $i$-th location point, typically represented as $p_i = (lon_i, lat_i)$, and $n$ represents the length of the trajectory.
\end{Def} 

\begin{Def}[Grid Partitioning]
    Given a geographical region bounded by latitude $[lat_{\min}, lat_{\max}]$ and longitude $[lon_{\min}, lon_{\max}]$, and the predefined cell size, the region is divided into a uniform grid of size $M \times U$, where $M$ and $U$ represent the number of divisions along the latitude and longitude axes, respectively. Each grid cell $g_{i,j}$ corresponds to a fixed rectangular subregion within the geographical area. Formally, the set of all grid cells in this spatial partition is defined as:
    $$
    \mathcal{G} = \{ g_{i,j} \mid 1 \leq i \leq M, \ 1 \leq j \leq U \},
    $$
    where $\mathcal{G}$ denotes the complete set of grid cells in the $M \times U$ partition.
    
\end{Def}

\begin{Def}[Trajectory Grid Sequence]
    A trajectory grid sequence, denoted as $\mathscr{T}_g$, is the sequence of grid cells obtained by mapping each point in a trajectory $\mathscr{T} = [p_1,p_2,\dots, p_n]$ to its corresponding grid cell in the grid set $\mathcal{G}$: 
    \[
    \mathscr{T}_{g} = [g_{i_1,j_1}, g_{i_2,j_2}, \ldots, g_{i_n,j_n}] \in \mathbb{R}^{n \times 2},
    \]
    where $g_{i_n,j_n}$ is the grid cell containing $p_n$.
\end{Def}

\begin{Def}[Heuristic Similarity]
    Heuristic measures quantify the similarity between two trajectories using predefined distance metrics, such as Hausdorff or Discrete Fr\'{e}chet distance. The computed similarity values are stored in a square matrix $\mathbf{H} \in \mathbb{R}^{N \times N}$, where $N$ represents the number of trajectories and each entry $\mathbf{H}_{ij}$ represents the distance between trajectories $\mathscr{T}_i$ and $\mathscr{T}_j$ under the chosen metric.
\end{Def}


\begin{Pro*}[Trajectory Similarity Computation]
Given a heuristic metric \( d(\cdot, \cdot) \) and two arbitrary trajectories \(\mathscr{T}_i \) and \( \mathscr{T}_j \), the objective is to learn a parameterized trajectory similarity function \( f_\theta(\cdot, \cdot) \) that minimizes the discrepancy between \( f_\theta(\mathscr{T}_i, \mathscr{T}_j) \) and \( d(\mathscr{T}_i, \mathscr{T}_j) \) as defined by the following equation:
\begin{equation}
\operatorname*{argmin}_\theta\ \left| f_\theta(\mathscr{T}_i, \mathscr{T}_j) - d(\mathscr{T}_i, \mathscr{T}_j) \right|.
\end{equation}
\end{Pro*}

In this work, we use DDBM to learn the reasonable evolution between any two trajectories, enhancing the model's robustness to noise. Next, we briefly review some basic concepts needed to understand diffusion models ~\cite{ddpm}.

\textbf{Diffusion process.} To model the data distribution $q_{\text{data}}(\mathbf{x})$ for $\mathbf{x} \in \mathbb{R}^d$, a stochastic diffusion process is introduced, defined by a sequence of time-indexed variables $\{\mathbf{x}_t\}_{t=0}^T$. This process begins with $\mathbf{x}_0 \sim p_0(\mathbf{x}):= q_{\text{data}}(\mathbf{x})$ and evolves such that the final state satisfies $\mathbf{x}_T \sim p_T(\mathbf{x}):= p_{\text{prior}}(\mathbf{x})$, where $p_{\text{prior}}(\mathbf{x})$ denotes a fixed prior distribution, such as the Gaussian distribution.
The forward dynamics of the system are governed by the following stochastic differential equation (SDE):
\begin{equation}
    d\mathbf{x}_t = \textbf{f}(\mathbf{x}_t, t)\,dt + g(t)\,d\mathbf{w}_t,
\label{eq:sed_forward}
\end{equation}
where $\mathbf{f}:\mathbb{R}^d \times [0, T] \rightarrow \mathbb{R}^d$ represents the drift function, $g(t)$ is a scalar diffusion coefficient, and $\mathbf{w}_t$ denotes a standard Wiener process. Under this formulation, the variable $\mathbf{x}_T$ follows the prior distribution. The time-reversed process that recovers $\mathbf{x}_0$ from $\mathbf{x}_T$ is characterized by:
\begin{equation}
    d\mathbf{x}_t = \mathbf{f}(\mathbf{x}_t, t) - g(t)^2 \nabla_{\mathbf{x}_t} \log p(\mathbf{x}_t)\,dt + g(t)\,d\mathbf{w}_t,
\end{equation}
where $p(\mathbf{x}_t)$ is the marginal density of $\mathbf{x}_t$ at time $t$. Additionally, this stochastic process admits an equivalent deterministic formulation known as the probability flow ODE~\cite{ODE}, which shares the same marginal distributions:
\begin{equation}
    d\mathbf{x}_t = \left[f(\mathbf{x}_t, t) - \frac{1}{2} g(t)^2 \nabla_{\mathbf{x}_t} \log p(\mathbf{x}_t)\right] dt.
\end{equation}
This formulation allows sampling from $q_{\text{data}}$ by first drawing $\mathbf{x}_T \sim q_{\text{data}}(y)$ and then solving the reverse-time SDE or ODE to obtain the corresponding $\mathbf{x}_0$.

\textbf{Denoising score matching.} The score function $\nabla_{\mathbf{x}_t} \log p(\mathbf{x}_t)$ can be estimated through a denoising score-matching objective, defined as:
\begin{equation}
    \mathcal{L}(\theta) = \mathbb{E} 
    \left[ \left\| s_\theta(\mathbf{x}_t, t) - \nabla_{\mathbf{x}_t} \log p(\mathbf{x}_t \mid \mathbf{x}_0) \right\|^2 \right],
\end{equation}
by minimizing this objective yields a score estimator $s_\theta^\ast(\mathbf{x}_t, t)$ that approximates the true conditional score. This formulation remains tractable due to the fact that the transition distribution $p(\mathbf{x}_t \mid \mathbf{x}_0)$ is explicitly defined as the Gaussian: $\mathbf{x}_t = \alpha_t \mathbf{x}_0 + \sigma_t \epsilon, \epsilon \sim \mathcal{N}(0, \mathbf{I})$, where $\alpha_t$ and $\sigma_t$ are time-dependent functions that control the scaling of the signal and the noise, respectively. In practice, it is common to express the diffusion process in terms of the signal-to-noise ratio (SNR), defined as $\alpha_t^2 / \sigma_t^2$.

\begin{table}[h!]
    \renewcommand{\arraystretch}{1.05}
    \centering
    \caption{Key notations and descriptions.}
    \begin{tabularx}{\linewidth}{c|X}
    \Xhline{2\arrayrulewidth}
        \textbf{Notations} & \textbf{Descriptions} \\
        \hline
        $\mathscr{T}$ & Trajectory data \\ 

        $\mathscr{T}^0$, $\mathscr{T}^T$ & Initial and target states of the DDBM during pre-training \\

        $\mathbf{e}_\text{gps}$, $\mathbf{e}_\text{grid}$ & Trajectory GPS feature and grid features \\

        $\mathscr{T}_g$ & Trajectory grid sequence \\
        
        $n$ & The number of points in a trajectory \\
        
        $N$ & Number of trajectories in dataset \\

        $d$ & Model embedding dimension \\

        $d_{\text{h}}$ & Hidden layer dimension \\
        
        $L$ & Total layers of Semantic Alignment Module \\
        
        $\mathcal{G}$ & The set of all spatial grid cells\\
        
        $\mathbf{H}$ & Heuristic similarity matrix \\

        $\mathbf{H}_\text{p}$ & Model predicted similarity \\

        $\bm{h}$ & Trajectory embedding \\
        
        $d(\cdot , \cdot)$ & Heuristic metric \\

        $\mathbf{PE}$ & Pre-encoding module \\
        
        $\lambda_q$, $\lambda_k$, $\lambda_v$ & Adaptive scaling factors \\

        $\gamma_1$, $\gamma_2$ & List-wise loss weighting coefficients \\
        
        $\epsilon$ & Feature fusion coefficient \\
        
        $\mathbf{Q}$, $\mathbf{K}$, $\mathbf{V}$ & Attention matrix \\
    \Xhline{2\arrayrulewidth}
    \end{tabularx}
    \label{tab:notations}
\end{table}

\section{Method}
Our proposed framework consists of three main components: (1) \textit{Semantic Alignment Module (SAM)}, (2) \textit{DDBM-based Noise-robust Pre-Training}, and (3) \textit{Overall Ranking-aware Regularization}. First, SAM bridges coarse and fine grained inputs by aligning their semantic meanings and integrating them into a unified representation. Second, DDBM-based noise-robust pre-training enables the model to learn a stochastic transformation between paired trajectories and improved its robustness to noise. Third, overall ranking-aware regularization incorporates global trajectory ranking information into the model’s optimization process, enabling it to better capture the overall ranking structure and improve performance on ranking-sensitive metrics.

\subsection{Semantic Alignment Module}
We propose SAM as a foundational architecture for trajectory similarity computation. It is specifically designed to explicitly model and align representations across sequences with heterogeneous granularities. Specifically, SAM comprises $L$ stacked Semantic Alignment Layer (SALayer), each layer consisting of two components: a dual semantic alignment attention module followed by a residual feedforward network. 
Specifically, given a trajectory $\mathscr{T}$, we first extract its fine-grained GPS feature $\mathbf{e}_\text{gps}\in \mathbb{R}^{n \times 2}$ and then project the points onto the corresponding grids to obtain the coarse-grained grid features $\mathbf{e}_\text{grid} \in \mathbb{R}^{n \times 2}$. Subsequently, we encode the inputs using the pre-encoding module $\mathbf{PE}$ to map them into a unified $d$-dimensional space. For SSPD and Hausdorff distance, $\mathbf{PE}$ is implemented as a linear layer, while for discrete Fréchet distance, it consists of a one-layer LSTM.The computation process is as follows:
\begin{equation}
    \mathbf{Z}^{\left(0\right)}_\text{gps} = \mathbf{PE}(\mathbf{e_\text{gps}}), \quad \mathbf{Z^{\left(0\right)}_\text{grid}} = \mathbf{PE}(\mathbf{e_\text{grid}}) 
\end{equation}
where $\mathbf{Z}^{\left(0\right)}_\text{gps},\mathbf{Z}^{\left(0\right)}_\text{grid}\in\mathbb{R}^{n \times d}$ and $d$ denotes the embedding dimension of the model. The pre-encoded representations $\mathbf{Z}^{\left(0\right)}_\text{gps}, \mathbf{Z}^{\left(0\right)}_\text{grid}$ are then progressively refined through the stacked SALayer. At each layer $l \in [1, L]$, the representations are updated as follows:
\begin{equation}
\mathbf{Z}^{\left(l\right)}_\text{gps}, \mathbf{Z}^{\left(l\right)}_\text{grid} = \text{SALayer}^l(\mathbf{Z}^{\left(l-1\right)}_\text{gps}, \mathbf{Z}^{\left(l-1\right)}_\text{grid}).
\end{equation}

In the following sections, we first introduce the Semantic Alignment Attention (SAA), and then describe the Dual SAA.

\subsubsection{Semantic Alignment Attention}
The core design of SAA lies in combines cross-attention mechanism with an adaptively scaled attention score mask strategy to bridge the semantic gap between inputs of two different scales. Given two input representations $\mathbf{Z}^{\left(0\right)}_\text{gps}, \mathbf{Z}^{\left(0\right)}_\text{grid} \in \mathbb{R}^{n \times d}$, we treat $\mathbf{Z}^{\left(0\right)}_\text{gps}$ as the query and $\mathbf{Z}^{\left(0\right)}_\text{grid}$ as the source for both key and value. Linear projections are then applied to obtain the query, key, and value matrices: $\mathbf{Q}, \mathbf{K}, \mathbf{V} \in \mathbb{R}^{n \times d}$.  The cross-attention score $\mathbf{A}_{\text{cross}}$ is computed via the scaled dot-product between $\mathbf{Q}$ and $\mathbf{K}$, while the self-attention score $\mathbf{A}_{\text{self}}$ is derived from the scaled dot-product between $\mathbf{K}$ and $\mathbf{V}$. These are formally defined as:
\begin{gather}
    \mathbf{Q} = \mathbf{Z}^{\left(0\right)}_\text{gps}\mathbf{W}^q,\mathbf{K} = \mathbf{Z}^{\left(0\right)}_\text{grid}\mathbf{W}^k, \mathbf{V} = \mathbf{Z}^{\left(0\right)}_\text{grid}\mathbf{W}^v,  \\
    \mathbf{A}_{\text{cross}} = \text{softmax} \left( \frac{\mathbf{QK}^\mathsf{T}}{\sqrt{d}} \right), \\
    \mathbf{A}_{\text{self}} \hspace{0.3em} =  \text{softmax} \left( \frac{\mathbf{KV}^\mathsf{T}}{\sqrt{d}} \right),
    \label{eqs1}
\end{gather}
where $\mathbf{W}^q, \mathbf{W}^k, \mathbf{W}^v \in \mathbb{R}^{d \times d}$ are learnable weight matrices. $\mathbf{A}_{\text{self}}$ captures intra-scale dependencies, and $\mathbf{A}_{\text{cross}}$ further facilitates cross-scale information exchange, progressively narrowing the semantic gap. To achieve more comprehensive alignment, additional mechanisms are required to fully bridge the representations across different granularities. To achieve this goal, we introduce an adaptive fusion mechanism that dynamically computes weighting coefficients for $\mathbf{A}_{\text{self}}$ and $\mathbf{A}_{\text{cross}}$ as follows: 
\begin{equation}
\begin{aligned}
\lambda_{\text{cross}} &= \exp\left( \sum_{i=1}^{d} \lambda_q^{(i)} \cdot \lambda_k^{(i)} \right),\\
\lambda_{\text{self}} &= \exp\left( \sum_{i=1}^{d} \lambda_k^{(i)} \cdot \lambda_v^{(i)} \right), \\
\mathbf{Z}^{\left(l\right)}_\text{gps} &= \text{SAA}(\mathbf{Z}^{\left(l-1\right)}_\text{gps}, \mathbf{Z}^{\left(l-1\right)}_\text{grid}) \\
             &= \left( \lambda_\text{self} \cdot \mathbf{A}_{\text{self}} 
             + \lambda_\text{cross} \cdot \mathbf{A}_{\text{cross}} \right) \mathbf{V},
\end{aligned}
\label{eqs2}
\end{equation}
where $\lambda_q$, $\lambda_k$, and $\lambda_v \in \mathbb{R}^d$ are learnable parameters that are used to derive the scaling factors $\lambda_{\text{self}}$ and $\lambda_{\text{cross}}$ for adaptive attention fusion. The output $\mathbf{Z}^{\left(l\right)}_\text{gps}$ is further updated through a feedforward network with residual connections, formulated as:
\begin{equation}
\mathbf{Z}^{\left(l\right)}_\text{gps} = \text{Norm} \left( \mathbf{Z}^{\left(l\right)}_\text{gps} + \text{FFN}(\mathbf{Z}^{\left(l\right)}_\text{gps}) \right),
\label{eqs3}
\end{equation}
where \(\text{Norm}(\cdot)\) denotes layer normalization.

\subsubsection{Dual Semantic Alignment Attention}

To further bridge the semantic gap between inputs of different granularities, we extend the SAA mechanism to a dual formulation. Specifically, after computing $\mathbf{Z}^{\left(l\right)}_\text{gps} = \text{SAA}(\mathbf{Z}^{\left(l-1\right)}_\text{gps}, \mathbf{Z}^{\left(l-1\right)}_\text{grid})$, we symmetrically reverse the roles of the two inputs by treating $\mathbf{Z}^{\left(l-1\right)}_\text{grid}$ as the query and $\mathbf{Z}^{\left(l-1\right)}_\text{gps}$ as the key-value source. Under this reversed configuration, we compute $\mathbf{Z}^{\left(l\right)}_\text{grid}$ via a symmetric attention operation similar to Eqs.~\eqref{eqs1},~\eqref{eqs2} and~\eqref{eqs3}.

After updating through \( L \) stacked SALayers, we obtain the final representations \( \mathbf{Z}^{\left(L\right)}_\text{gps} \) and \( \mathbf{Z}^{\left(L\right)}_\text{grid} \). These two representations are then fused into a unified embedding via a weighted combination controlled by a hyperparameter $\epsilon$:
\begin{equation}
\bm{h} = \epsilon \cdot \mathbf{Z}^{\left(L\right)}_\text{gps} + (1 - \epsilon) \cdot \mathbf{Z}^{\left(L\right)}_\text{grid},
\end{equation}
where $\mu \in (0, 1)$ serves as a balancing hyperparameter that controls the relative contribution of $\mathbf{Z}^{\left(L\right)}_\text{gps}$ and $\mathbf{Z}^{\left(L\right)}_\text{grid}$ to the final fused representation $\bm{h}$. 

In summary, the proposed Dual SAA significantly enhances the model's ability to capture nuanced cross-granular interactions, leading to more comprehensive and expressive representations in practice.

\subsection{DDBM-based Noise-robust Pre-Training}
As discussed above, trajectory data is often affected by noise. To address this issue, we adopt DDBM as the pretraining strategy, which enables the model to learn interactions and transfer patterns between arbitrary pairs of trajectories, enhancing robustness and generalization. We first introduce DDBM and then explain how it is adapted for our task. Figure~\ref{fig:pretrain} presents an overview of the pretraining process.

\subsubsection{Denoising Diffusion Bridge Models}
The objective of DDBM is to learn bidirectional conditional generation directly from the initial state $\mathbf{x}_0$ to the terminal state $\mathbf{x}_T$ through the diffusion bridge. In the forward process, DDBM constructs a bridge diffusion process with fixed endpoints, transporting the initial state \( \mathbf{x}_0 \) to the target state \( \mathbf{x}_T = y \). This process is formulated by modifying the drift term using Doob's $h$-transform~\cite{doob1,doob2}. Accordingly, Eq.\eqref{eq:sed_forward} is rewritten as follows:
\begin{equation}
    d\mathbf{x}_t = \mathbf{f}(\mathbf{x}_t, t) dt + {g(t)^2}( \mathbf{h}_t )dt + g(t) d\mathbf{w}_t,
\end{equation}
where $\mathbf{h}_t=\mathbf{h}(x, t, y, T) = \nabla_{\mathbf{x}_t} \log p(\mathbf{x}_T=y| \mathbf{x}_t=x)$ denotes the gradient of the log transition kernel, guaranteeing almost sure convergence to \( y \) at terminal time \( T \). Using this equation, we can move from $\mathbf{x}_0$ to $\mathbf{x}_T$.

In the reverse process, to sample from the terminal state $\mathbf{x}_T$ back to $\mathbf{x}_0$, DDBM formulates the time‐reversed dynamics of the conditional distribution $q(\mathbf{x}_t \mid \mathbf{x}_T)$:
\begin{equation}
    d\mathbf{x}_t = \left[\mathbf{f}(\mathbf{x}_t, t) - g(t)^2 (\mathbf{s}_t - \mathbf{h}_t) \right] dt + g(t)\, d\hat{\mathbf{w}}_t, 
\end{equation}
where $\mathbf{s}_t=\mathbf{s}(x, t, y, T) = \nabla_{\mathbf{x}_t} \log q(\mathbf{x}_t=x \mid \mathbf{x}_T=y)$ is the gradient of the conditional probability distribution of the state \( x_t \) given the terminal state \( x_T \) which is learned by neural network, and $\hat{\mathbf{w}}_t$ denotes the reverse‐time Wiener process.

For loss computation, since the true bridge distribution score \(\mathbf{s}(x, t, y, T)\) is not available in closed form, DDBM leverages the tractable Gaussian bridge marginal \(q(\mathbf{x}_t \mid \mathbf{x}_0, \mathbf{x}_T)\) to define the training objective:
\begin{equation}
    \mathcal{L}(\theta)
      = \mathbb{E}_{\mathbf{x}_0, \mathbf{x}_T, t ,\mathbf{x}_t}
      \bigl\lVert\,s_\theta(\mathbf{x}_t, \mathbf{x}_T, t) \;-\; \nabla_{\mathbf{x}_t}\log q(\mathbf{x}_t\mid \mathbf{x}_0, \mathbf{x}_T)\bigr\rVert^2,
\end{equation}
minimizing this loss drives the network output \(s_\theta\) to match the true conditional score \(\nabla_{\mathbf{x}_t}\log q(\mathbf{x}_t\mid \mathbf{x}_T)\). In practice, DDBM adopts the pred-$\mathbf{x}$ parameterization from EDM~\cite{karras2022elucidating}, We also apply this technique, and redefine the loss calculation as the MSE loss between the true value $\mathbf{x}$ and the model output. For more details, please refer to DDBM~\cite{ddbm}.

\subsubsection{DDBM Pre-training}
Given an arbitrary pair of distinct trajectories, one is designated as the initial state $\mathscr{T}^0$, while the other serves as the terminal state $\mathscr{T}^T$ at the final time step. We extract their GPS and grid features at both the initial and final states, \( \mathbf{e}^0_{\mathrm{gps}}, \mathbf{e}^0_{\mathrm{grid}} \) and \( \mathbf{e}^T_{\mathrm{gps}}, \mathbf{e}^T_{\mathrm{grid}} \), respectively. Subsequently, following the DDBM sampling process, we obtain the intermediate time data $\mathbf{e}^t_\text{gps}$ at time $t$ by sampling from $\mathbf{e}^0_{\text{gps}}$ and $\mathbf{e}^T_{\text{gps}}$, the formula is:
\begin{equation}
\begin{aligned}
q(\mathbf{x}_t \mid \mathbf{x}_0, \mathbf{x}_T) &= \mathcal{N}\bigl(\hat\mu_t,\,\hat\sigma_t^2 I\bigr), \\[6pt]
\hat\mu_t &= \frac{\mathrm{SNR}_T}{\mathrm{SNR}_t}\,\frac{\alpha_t}{\alpha_T}\,\mathbf{e}^T_\text{gps} \;+\;\alpha_t\,\mathbf{e}^0_\text{gps}\Bigl(1 - \frac{\mathrm{SNR}_T}{\mathrm{SNR}_t}\Bigr), \\[6pt]
\hat\sigma_t^2 &= \sigma_t^2\Bigl(1 - \frac{\mathrm{SNR}_T}{\mathrm{SNR}_t}\Bigr), \\
\mathbf{e}^t_\text{gps} &=  \hat{\mu}_t + \hat{\sigma}_t \times \mathbf{e}^t_\text{Noise},
\end{aligned}
\label{eq:sample}
\end{equation}
where \( \hat{\mu}_t \) denotes the ground truth of the sample \( \mathbf{e}^t_{\mathrm{gps}} \), \( \hat{\sigma}_t^2 \) represents the corresponding noise variance and \( \mathbf{e}^t_{\text{Noise}} \) denotes randomly Gaussian noise. In addition, \( \alpha_t \) and \( \sigma_t \) are the pre-defined signal and noise schedules, respectively, and the signal-to-noise ratio (SNR) is computed as \( \mathrm{SNR}_t = \alpha_t^2 / \sigma_t^2 \). In practice, we adopt a linear noise schedule defined as \( \beta(t) = \beta_{\text{min}} + t(\beta_{\text{max}} - \beta_{\text{min}}) \), which yields the signal coefficient:
\begin{equation}
    \alpha_t = \exp\left(-\frac{1}{2} \int_0^t \beta(s) \, ds \right).
\end{equation}

Using Eq.~\eqref{eq:sample}, we obtain the intermediate state $\mathbf{e}^t_{\text{gps}}$, which serves as one of the dual-scale inputs to the model. Since $\mathbf{e}^t_{\text{gps}}$ contains only fine-grained and potentially noisy information, we further compute a noise-free, coarse-grained representation $\mathbf{e}^t_\text{grid}$ from the grid view to complement the input. 
The formula is given as follows:
\begin{equation}
    \mathbf{e}^t_\text{grid}=\frac{\mathrm{SNR}_T}{\mathrm{SNR}_t}\,\frac{\alpha_t}{\alpha_T}\,\mathbf{e}^T_\text{grid} \;+\;\alpha_t\,\mathbf{e}^0_\text{grid}\Bigl(1 - \frac{\mathrm{SNR}_T}{\mathrm{SNR}_t}\Bigr),
\end{equation}
this auxiliary feature $\mathbf{e}^t_\text{grid}$ is incorporated into the SAM with $\mathbf{e}^t_{\text{gps}}$ to provide stable structural guidance and help the model better capture the high-level semantic patterns of the trajectory.

Then, we can obtain the trajectory embedding using the following formula:
\begin{equation}
    \begin{gathered}
        \mathbf{Z}^t_\text{gps} = \mathbf{PE}(\mathbf{e}^t_\text{gps}), \quad \mathbf{Z}^t_\text{grid} = \mathbf{PE}(\mathbf{e}^t_\text{grid}), \\
        \bm{h} = \text{SAM}\left(\mathbf{Z}^t_\text{gps} , \mathbf{Z}^t_\text{grid} \right),
    \end{gathered}
\end{equation}
where \( \bm{h} \in \mathbb{R}^{n \times d} \) denotes the reconstruction of \( \hat{\mu} \) in the embedding space. We follow the pred-\( \mathbf{x} \) parameterization for loss computation, while computing the loss in the embedding space rather than the original data space. The objective is formulated as:
\begin{equation}
    \begin{gathered}
        \bm{h}_{\hat{\mu}} = \mathbf{PE}(\hat{\mu}),\\
        \mathcal{L}_{\text{MSE}} = \left\| \bm{h} - \bm{h}_{\hat{\bm{\mu}}} \right\|^2.
    \end{gathered}
\end{equation}

By computing the loss in the high-dimensional space, we can better optimize the model's performance in the feature space, improving the accuracy and stability of trajectory similarity computation, while also mitigating the influence of local noise in the original data space to enable the model to capture the global similarity between trajectories more accurately.



\subsection{Overall Ranking-aware Regularization}
To explore the global trajectory-level ranking information that has been overlooked in previous studies, we introduce the overall ranking-aware regularization, which combines two list-wise ranking objectives: \textbf{ListNet}\cite{listnet} and \textbf{Rank-Decay ListNet}. We first provide a detailed explanation of how ListNet and Rank-Decay ListNet are utilized, along with the overall loss function employed during the fine-tuning stage.

\subsubsection{ListNet}
Let ${s} = [s_1, s_2, \dots, s_k]$ denote the predicted similarity scores for $k$ candidate trajectories with respect to a query trajectory, and let ${r} = [r_1, r_2, \dots, r_k]$ denote the corresponding heuristic similarity labels. We adopt the list-wise learning-to-rank paradigm, which treats the entire set of candidates as a single learning instance, rather than focusing on individual items or item pairs.

Specifically, we adopt the top-one probability formulation proposed in ListNet, which models the probability of each candidate being ranked at the top based on its similarity score. Given a list of predicted similarity scores ${s}$ and the corresponding ground-truth scores ${r}$, the top-one probability distributions are computed using the softmax function:
\begin{align}
P(i \mid {s}) &= \frac{\exp(s_i)}{\sum_{j=1}^k \exp(s_j)}, \\
P(i \mid {r}) &= \frac{\exp(r_i)}{\sum_{j=1}^k \exp(r_j)},
\end{align}
where $P(i \mid \mathbf{s})$ denotes the model-predicted probability of the $i$-th candidate being ranked at the top, and $P(i \mid \mathbf{r})$ denotes the target top-one probability for the $i$-th candidate, computed from the heuristic similarity labels $\mathbf{r}$. The ListNet loss is then defined as the cross-entropy between the target and predicted top-one probability distributions:
\begin{equation}
\mathcal{L}_{\text{ListNet}} = -\sum_{i=1}^k P(i \mid {r}) \log P(i \mid {s}).
\end{equation}

Since the ground truth distribution $P(\cdot \mid {r})$ is fixed, its entropy does not affect the optimization objective. Therefore, minimizing the cross-entropy is equivalent to making the model-predicted ranking distribution $P(\cdot \mid {s})$ closer to the reference distribution $P(\cdot \mid {r})$, enabling the model to better learn the correct ranking relationships.

\subsubsection{Rank-Decay ListNet}
While ListNet leverages top-one probability distributions derived from similarity scores to learn list-wise ranking preferences, it implicitly treats all candidates equally, regardless of their importance in the final ranking. However, in practical retrieval scenarios, ranking errors among the top candidates are often more critical than those among lower-ranked ones. To better reflect this intuition during training, we introduce Rank-Decay ListNet (RD-ListNet), which incorporates rank-aware weights into the ListNet loss.

Specifically, we first obtain a relevance-based permutation $\mathbf{r}^\prime = [r^\prime_1, r^\prime_2, \dots, r^\prime_k]$ by sorting the ground-truth similarity scores $\mathbf{r}$ in descending order. The corresponding rank-aware decay weights are then computed using a logarithmic function:
\begin{equation}
w_i = \frac{1}{\log_2(i + 1)},
\end{equation}
where $i$ denotes the position index in the sorted label sequence, with smaller $i$ indicating higher relevance. These decay weights are used to scale the cross-entropy loss between the predicted and ground-truth top-one distributions:
\begin{equation}
\mathcal{L}_{\text{RD-ListNet}} = - \sum_{i=1}^k w_i \cdot P(i \mid \mathbf{r}) \log P(i \mid \mathbf{s}),
\end{equation}
where $P(i \mid {s})$ and $P(i \mid \mathbf{r})$ are the softmax-normalized similarity scores and heuristic labels, respectively. By assigning higher weights to top-ranked items, RD-ListNet encourages the model to focus more on accurately ranking the most relevant trajectories.




\subsubsection{Total Loss}

To jointly leverage both point-wise and list-wise supervision signals, we combine the mean squared error loss with overall ranking-aware regularization. The overall training loss during fine-tuning is defined as:
\begin{equation}
\mathcal{L}_{\text{total}} = \mathcal{L}_{\text{MSE}} + \gamma_1 \mathcal{L}_{\text{ListNet}} + \gamma_2 \mathcal{L}_{\text{RD-ListNet}},
\end{equation}
where $\gamma_1$ and $\gamma_2$ are weighting hyperparameters that balance the contributions of the two list-wise components during training. We adopt this joint loss formulation during the fine-tuning stage, enabling the model to refine its similarity predictions with respect to both absolute magnitude and relative order. In our experiments, we find that this hybrid optimization strategy leads to improved generalization and more stable query performance, particularly in noisy or ambiguous scenarios where point-wise supervision alone may be inadequate.

\subsection{Model Fine-Tuning}
We define the fine-tuning pseudo-code as Algorithm~\ref{alg:TrajDiff}, where dual-scale features are iteratively aligned and fused through $L$ stacked SALayers, followed by supervised optimization with total loss. Specifically, given two inputs from different scales $\mathbf{e}_{\text{gps}}$ and $\mathbf{e}_{\text{grid}}$, we first project them into a unified embedding space using the pre-encoding module $\mathbf{PE}$, as shown in Line 2, obtaining initial representations $\mathbf{Z}^{\left(0\right)}_\text{gps}$ and $\mathbf{Z}^{\left(0\right)}_\text{grid}$. These embeddings are then passed through $L$ layers of SALayer, where Dual SAA is performed iteratively to produce the final representations $\mathbf{Z}^{\left(L\right)}_\text{gps}$ and $\mathbf{Z}^{\left(L\right)}_\text{grid}$, as illustrated in Lines 3–5. Subsequently, these two outputs are fused using a hyperparameter $\epsilon$ and aggregated via mean pooling to obtain the final embedding $\bm{h}$. Then $\bm{h}$ is used to compute the predicted similarity matrix $\mathbf{H}_\text{p}$, which is then compared against the ground truth $\mathbf{H}$ to calculate the total loss, as defined in Line 9. Finally, the model is updated via backpropagation.

\begin{algorithm}[H]
  \caption{The Fine-Tuning Process of \model}
  \label{alg:TrajDiff}
  \begin{algorithmic}[1]
  \renewcommand{\algorithmicrequire}{\textbf{Input}}
  \renewcommand{\algorithmicensure}{\textbf{Output}}
    \REQUIRE
      Different scale representations $\mathbf{e}_\text{gps}$ and $\mathbf{e}_\text{grid}$, per-encoding module $\mathbf{PE}$, ground truth $\mathbf{H}$, pretrained SAM
    \ENSURE
      Embedding results $\bm{h}$
    \STATE \textbf{Aggregate two scale inputs using the SAM;}
    \STATE Project the initial input\\
    $\mathbf{Z}_\text{gps}^{\left(0\right)} \gets pe(\mathbf{z}_\text{gps})$, $\mathbf{Z}^{\left(0\right)}_\text{grid} \gets pe(\mathbf{z}_\text{grid})$;
    \FOR{$l \in [1, L]$}   
        
        \STATE Duplicate $\mathbf{Z}^{\left(l-1\right)}_\text{gps}$ and $\mathbf{Z}^{\left(l-1 \right)}_\text{grid}$\\
        $\mathbf{Z}_{\text{gps}}^{\left(l-1\right)}{}^{\prime} \gets \mathbf{Z}^{\left(l-1\right)}_\text{gps}$, $\mathbf{Z}_{\text{grid}}^{\left(l-1\right)}{}^{\prime} \gets \mathbf{Z}^{\left(l-1\right)}_\text{grid}$;
        
        \STATE Execute the Dual Aggregation Attention\\
        $\mathbf{Z}^{\left(l\right)}_\text{gps} \gets \text{SAA}\left(\mathbf{Z}^{\left(l-1\right)}_\text{gps},\mathbf{Z}_{\text{grid}}^{\left(l-1\right)}{}^{\prime} \right)$\\
        $\mathbf{Z}^{\left(l\right)}_\text{grid} \gets \text{SAA}\left(\mathbf{Z}^{\left(l-1\right)}_\text{grid},\mathbf{Z}_{\text{gps}}^{\left(l-1\right)}{}^{\prime} \right)$;
    \ENDFOR
    
    \STATE Fuse two embeddings\\
    $\bm{h} \gets \epsilon \mathbf{Z}^{\left(L\right)}_\text{gps} + \left(1 - \epsilon \right)\mathbf{Z}^{\left(L\right)}_\text{grid}$\\
    $\bm{h} \gets mean\_pooling\left(\bm{h}\right)$;

    \STATE Calculate embedding distance $\mathbf{H}_\text{p}$ from $\bm{h}$;\\
    
    \STATE Calculate total loss using $\mathbf{H}_\text{p}$ and $\mathbf{H}$\\
    $\mathcal{L} \gets \mathcal{L}_{MSE}  + \gamma_1 \mathcal{L}_{ListNet} + \gamma_2 \mathcal{L}_{RD-ListNet}$;
    
    \STATE Backpropagation and update parameters in \model;
    \RETURN $\bm{h}$
  \end{algorithmic}
\end{algorithm}

\section{Experiments} 
\label{sec:experiments}

In this section, we first present the experimental setup, including the datasets, baseline methods, parameter configurations, evaluation metrics, and implementation environment. We then conduct extensive experiments to evaluate the effectiveness of our proposed model, \model, aiming to address the following research questions:

\begin{itemize}
    \item \textbf{RQ1:} Can \model achieves superior performance in trajectory similarity computation compared to several state-of-the-art baselines?
    \item \textbf{RQ2:} How does each module of \model contribute to its overall performance?
    \item \textbf{RQ3:} What is the impact of the major hyperparameters of \model on the similarity computation?
    \item \textbf{RQ4:} How does \model perform in terms of efficiency?
\end{itemize}

\subsection{Experimental Settings}
\subsubsection{Datasets}
We evaluate our method on three publicly available datasets: 

(1) \textbf{Proto}\footnote{\href{https://www.kaggle.com/c/pkdd-15-predict-taxi-service-trajectory-i/data}{https://www.kaggle.com/c/pkdd-15-predict-taxi-service-trajectory-i/data}} contains 1.7 million taxi trajectories from Porto, Portugal (July 2013 – June 2014); 

(2) \textbf{Geolife}\footnote{\href{https://www.microsoft.com/en-us/research/publication/geolife-gps-trajectory-dataset-user-guide/}{https://www.microsoft.com/en-us/research/publication/geolife-gps-trajectory-dataset-user-guide/}}~\cite{geolife} includes trajectories of 182 users in Beijing (April 2007 – August 2012); 

(3) \textbf{T-Drive}\footnote{\href{https://www.microsoft.com/en-us/research/publication/t-drive-trajectory-data-sample/}{https://www.microsoft.com/en-us/research/publication/t-drive-trajectory-data-sample/}}~\cite{tdriver1} provides GPS trajectories of 10,357 taxis in Beijing from February 2 to 8, 2008. 

In line with prior work~\cite{trajcl}, we remove trajectories outside the geographic boundaries and those that do not meet the required length criteria. Since both Geolife and T-Drive are based in Beijing, we apply the same geographic boundary for consistency.

For T-Drive, since each trajectory contains data from many days, we segment multi-day taxi trajectories into sub-trajectories by identifying stay points, which are locations where a vehicle remains within a 100-meter radius for over 5 minutes. Additional details of the datasets are summarized in Tab.~\ref{table:dataset_statistics}. 

For the Porto dataset, we select the first 200,000 trajectories for pre-training and additionally sample 10,000 trajectories for similarity computation. For both the T-Drive and Geolife datasets, we randomly sample 10,000 trajectories to compute pairwise similarities. Due to the smaller size of the Geolife and T-Drive datasets, we fine-tune the model pre-trained on the Porto dataset separately on these two datasets. The trajectories used for similarity computation are split into training, evaluation, and test sets with a ratio of 7:1:2.

\begin{table}[h!]
\centering
\caption{Detailed statistics of the dataset.}
\resizebox{\linewidth}{!}{
\begin{tabular}{cccc}
\hline
\textbf{Dataset}                 & \textbf{Porto}       & \textbf{Geolife}    & \textbf{T-Driver} \\ \hline
Trajectories (\(n\))           &   1,372,725          & 10,940             & 28,843                  \\
min length        & 20                   & 20                 & 20                      \\
max length        & 200                  & 300                & 200                     \\
mean length       &  48                  & 106                &38                      \\
longitude        & [-8.519, -8.005]       & [116.25, 116.5]       & [116.25, 116.5]        \\
latitude         & [41.1001, 41.2086]      & [39.8, 40.1]          & [39.8, 40.1]           \\ \hline
\end{tabular}
}
\label{table:dataset_statistics}
\end{table}

\subsubsection{Baseline}
We focus primarily on trajectory similarity computation with the goal of more accurately approximating heuristic measures. To validate the effectiveness of our model, we compare \model with eight relevant methods.

\begin{itemize}
    \item \textbf{t2vec~\cite{t2vec}}: It learns trajectory embeddings via seq2seq modeling, enabling fast and noise-robust similarity computation.
    \item \textbf{NeuTraj~\cite{neutraj}}: It is a neural metric learning model that approximates trajectory similarity using spatial attention and seed-guided supervision in linear time.
    \item \textbf{Traj2SimVec~\cite{zhang2020trajectory}}: It learns trajectory similarity through deep representation learning with auxiliary supervision, leveraging sub-trajectory distance loss and point matching loss for improved robustness and scalability.
    \item \textbf{T3S~\cite{T3S}}: It learns trajectory embeddings by combining LSTM and self-attention to capture spatial and structural features for similarity computation.
    \item \textbf{TMN~\cite{tmn}}: It approximates trajectory similarity via attention-based point matching between trajectory pairs.
    \item \textbf{TrajGAT~\cite{trajgat}}: It models trajectory similarity using a graph-based Transformer with hierarchical spatial encoding for long-range dependency capture.
    \item \textbf{TrajCL~\cite{trajcl}}: It employs contrastive learning with dual-feature self-attention and trajectory augmentations to efficiently learn unsupervised trajectory similarity representations with strong generalization.
    \item \textbf{KGTS~\cite{chen2024kgts}}: It proposes a knowledge-guided spatiotemporal graph neural network for traffic forecasting, which enhances prediction accuracy by incorporating external knowledge to guide the construction and modeling of graph structures, thereby capturing complex spatiotemporal dependencies more effectively.

\end{itemize}

\subsubsection{Parameter Settings}
We set the batch size to 128, the learning rate to 0.001, and the number of attention heads to 16. The experimental results were obtained using random seeds 18, 66, and 108, with each experiment conducted three times and the average value taken. During fine-tuning, we set $\gamma_1 = 0.1$ and $\gamma_2 = 0.001$, and the number of SAM layers is set to 1. We pretrain for 20 epochs with early stopping patience of 5 epochs, and fine-tune for 30 epochs with early stopping patience of 10 epochs.


\subsubsection{Evaluation Metrics and Environment Implementation}
Following previous works~\cite{trajcl}, we adopt the top-$k$ hit ratio (HR@$k$) and top-$k$ recall for top-$t$ ground truth (Recall-$t$@$k$) as our evaluation metrics. Both metrics assess the effectiveness of models in top-$k$ similarity search tasks. Specifically, HR@$k$ measures the overlap between the top-$k$ trajectories retrieved using approximate similarity and those retrieved using ground-truth similarity. Recall-$t$@$k$, on the other hand, evaluates the model’s ability to retrieve the top-$t$ ground-truth items within the top-$k$ results. In this paper, we primarily report HR@1, HR@5, HR@20, and H5@20. Experiments are conducted on dual-socket AMD EPYC 7402 CPUs (2 × 24 cores, 96 threads) @ 2.80GHz and a GeForce RTX 4090 GPU.

\begin{table*}[!htb]
\renewcommand{\arraystretch}{1.13}
\caption{Performance comparison of the proposed model and different baselines. Marker * indicates the results are statistically significant (t-test with p-value $<$ 0.01).}
\label{tab:results}
\resizebox{\textwidth}{!}{
\begin{tabular}{cc|cccc|cccc|cccc}
\Xhline{1pt}
\multirow{2}{*}{Dataset} & \multirow{2}{*}{Method} & \multicolumn{4}{c}{SSPD} & \multicolumn{4}{c}{Discrete Fre\'chet} & \multicolumn{4}{c}{Hausdorff} \\ \cline{3-14} 
 &  & HR@1 & HR@5 & HR@20 & R5@20 & HR@1 & HR@5 & HR@20 & R5@20 & HR@1 & HR@5 & HR@20 & R5@20 \\ \hline
\multirow{8}{*}{Porto} 
 & t2vec & 0.271 & 0.348 & 0.504 & 0.675 & 0.467 & 0.519 & 0.667 & 0.889 & 0.359 & 0.410 & 0.564 & 0.793 \\
 & Traj2SimVec & 0.313 & 0.401 & 0.487 & 0.723 & 0.471 & 0.524 & 0.671 & 0.893   & 0.301 & 0.342 & 0.423 & 0.551 \\
 & T3S & 0.288 & 0.412 & 0.534 & 0.763 & 0.461 & 0.581 & 0.723 & 0.946 & 0.490 & 0.610 & 0.691 & 0.941 \\
 & TMN & 0.323 & 0.413 & 0.497 & 0.746 & 0.473 & 0.568 & 0.656 & 0.925 & 0.481 & 0.573 & 0.661 & 0.893 \\
 & TrajGAT & 0.331 & 0.426 & 0.590 & 0.691 & 0.330 & 0.352 & 0.399 & 0.694      & \underline{0.568} & \underline{0.683} & \underline{0.739} & \underline{0.961} \\
 & TrajCL & \underline{0.376} & \underline{0.495} & \underline{0.583} & \underline{0.821}  & \underline{0.493} & \underline{0.618} & \underline{0.740} & \underline{0.955} & 0.516 & 0.645 & 0.721 & 0.955 \\
 & KGTS & 0.301 & 0.369 & 0.537 & 0.680 & 0.436 & 0.531 & 0.658 & 0.876 & 0.431 & 0.528 & 0.607 & 0.817 \\
 & \textbf{Ours} & \textbf{0.485*} & \textbf{0.573*} & \textbf{0.657*} & \textbf{0.894*} & \textbf{0.541*} & \textbf{0.664*} & \textbf{0.762*} & \textbf{0.971*} & \textbf{0.583*} & \textbf{0.685*} & \textbf{0.747*} & \textbf{0.969*} \\ \hline
 & \textit{Improvement} & 28.98\% & 15.76\% & 12.69\% & 8.89\% & 9.74\% & 7.44\% & 2.97\% & 1.68\%  & 2.64\% & 0.29\% & 1.08\% & 0.83\% \\ \hline

\multirow{8}{*}{Geolife} 
 & t2vec & 0.240 & 0.378 & 0.513 & 0.707 & 0.402 & 0.451 & 0.601 & 0.782 & 0.302 & 0.360 & 0.521 & 0.736 \\
 & Traj2SimVec & 0.264 & 0.398 & 0.542 & 0.749 & 0.432 & 0.482 & 0.592 & 0.772   & 0.254 & 0.302 & 0.383 & 0.511 \\
 & T3S & 0.254 & 0.401 & 0.557 & 0.773 & 0.342 & 0.482 & 0.625 & 0.802 & 0.244 & 0.322 & 0.444 & 0.633 \\
 & TMN & 0.268 & 0.399 & 0.546 & 0.769 & 0.385 & 0.501 & 0.624 & 0.794 & 0.373 & 0.523 & 0.681 & 0.845 \\
 & TrajGAT & 0.281 & 0.413 & 0.554 & \underline{0.796} & 0.392 & 0.531 & 0.712 & 0.901 & \underline{0.418} & 0.584 & 0.734 & 0.904 \\
 & TrajCL & \underline{0.301} & \underline{0.443} & \underline{0.573} & 0.792    & \underline{0.400} & \underline{0.526} & \underline{0.704} & \underline{0.891} & 0.410 & \underline{0.587} & \underline{0.736} & \underline{0.906} \\
 & KGTS & 0.273 & 0.405 & 0.521 & 0.763 & 0.391 & 0.468 & 0.631 & 0.803 & 0.334 & 0.381 & 0.549 & 0.729 \\
 & \textbf{Ours} & \textbf{0.420*} & \textbf{0.541*} & \textbf{0.651*} & \textbf{0.847*} & \textbf{0.507*} & \textbf{0.646*} & \textbf{0.742*} & \textbf{0.929*} & \textbf{0.579*} & \textbf{0.693*} & \textbf{0.778*} & \textbf{0.953*} \\ \hline
 & \textit{Improvement}  & 39.53\% & 22.12\% & 13.61\% & 6.40\%  & 26.75\% & 22.81\% & 5.40\% & 4.26\%  & 38.52\% & 18.06\% & 5.71\% & 5.19\% \\ \hline

\multirow{8}{*}{T-Driver} 
 & t2vec & 0.137 & 0.278 & 0.414 & 0.607 & 0.193 & 0.381 & 0.401 & 0.702 & 0.214 & 0.303 & 0.485 & 0.709 \\
 & Traj2SimVec   & 0.139 & 0.284 & 0.426 & 0.611 & 0.244 & 0.432 & 0.525 & 0.747 & 0.254 & 0.302 & 0.383 & 0.511 \\
 & T3S & 0.134 & 0.265 & 0.411 & 0.567 & 0.222 & 0.401 & 0.497 & 0.732 & 0.324 & 0.472 & 0.621 & 0.833 \\
 & TMN & 0.148 & 0.281 & 0.426 & 0.609 & 0.276 & 0.441 & 0.547 & 0.794 & 0.325 & 0.489 & 0.631 & 0.845 \\
 & TrajGAT & 0.161 & 0.296 & 0.431 & 0.616 & 0.296 & 0.451 & 0.579 & \underline{0.829} & 0.330 & 0.519 & 0.649 & 0.902 \\
 & TrajCL & \underline{0.167} & \underline{0.302} & \underline{0.449} & \underline{0.629} & \underline{0.307} & \underline{0.464} & \underline{0.581} & 0.826 & \underline{0.337} & \underline{0.524} & \underline{0.651} & \underline{0.906} \\
 & KGTS & 0.143 & 0.294 & 0.437 & 0.619 & 0.286 & 0.425 & 0.553 & 0.812 & 0.301 & 0.4571 & 0.609 & 0.889 \\
 & \textbf{Ours} & \textbf{0.321*} & \textbf{0.461*} & \textbf{0.577*} & \textbf{0.794*} & \textbf{0.406*} & \textbf{0.542*} & \textbf{0.660*} & \textbf{0.912*} & \textbf{0.438*} & \textbf{0.591*} & \textbf{0.699*} & \textbf{0.933*} \\ \hline
 & \textit{Improvement} & 92.22\% & 52.64\% & 28.51\% & 26.23\% & 32.25\% & 16.81\% & 13.60\% & 10.01\% & 29.97\% & 12.79\% & 7.37\% & 2.98\% \\ \Xhline{1pt}
\end{tabular}
}
\end{table*}

\subsection{Performance Comparison (RQ1)}\label{RQ1}
Table~\ref{tab:results} presents a comprehensive performance comparison across various methods on three benchmark datasets. Our proposed model, \model, consistently achieves the best performance. Specifically, the average improvements in HR@1 across the three heuristic measure are 53.57\%, 22.91\%, and 23.71\%, respectively, demonstrating the superiority of the \model in trajectory similarity computation.

First, existing studies (e.g., TrajCL, KGTS, TrajGAT, and T3S) primarily focus on single-scale trajectory representations, such as GPS-based features or grid-based features. In addition, these methods often rely on simplistic feature fusion strategies, including direct embedding summation or independent attention score masking. Such designs limit the ability of the model to fully exploit the complementary information inherent in dual-scale trajectory data, thereby resulting in suboptimal performance. In contrast, our SAM explicitly models the correlations between dual-scale features to achieve effective semantic alignment, leading to more expressive and discriminative trajectory representations. As shown in Table~\ref{tab:results}, our method consistently outperforms the aforementioned baselines across all heuristic metric. To further verify the effectiveness of the SAM, we replace it with the fusion strategies employed in TrajCL and T3S and evaluate the performance on the Porto dataset using three heuristic metric. We summarize the results in Table~\ref{tab:fusion-method}, indicating that substituting our SAM with the fusion strategies from TrajCL and T3S consistently leads to performance degradation across all three heuristic metrics. Specifically, the HR@1 scores exhibit average decreases of 50.32\%, 40.52\%, and 27.3\% under three heuristic metrics. These findings demonstrate that the proposed SAM effectively mitigates semantic inconsistency and enhances the integration of dual-scale trajectory features.

Second, we pretrain our model using the denoising diffusion bridge, which enables it to learn a stochastic transformation between paired trajectories and enhances its robustness to noise. Importantly, instead of computing loss in the raw data space, we optimize the model in a high-dimensional embedding space. This allows the model to more effectively suppress local noise while preserving global semantic structure, resulting in more accurate and stable representations. As shown in Table~\ref{tab:results}, our method achieves superior performance, particularly on high-noise datasets such as T-Driver and GeoLife. In contrast, on the Porto dataset where the trajectory data is relatively clean, the performance gain is moderate, reflecting the reduced necessity for denoising. This variation across datasets highlights the effectiveness of our denoising strategy in recovering informative representations from corrupted inputs. Overall, these findings confirm the advantage of incorporating a denoising diffusion bridge during pretraining and optimizing in the embedding space, equipping the model with improved resilience and generalization under noisy or ambiguous conditions.

\begin{figure}[thbp!]
    \centering
    \begin{minipage}[t]{1.0\linewidth}
    \centering
        \begin{tabular}{@{\extracolsep{\fill}}c@{}@{\extracolsep{\fill}}}
            \includegraphics[width=\linewidth]{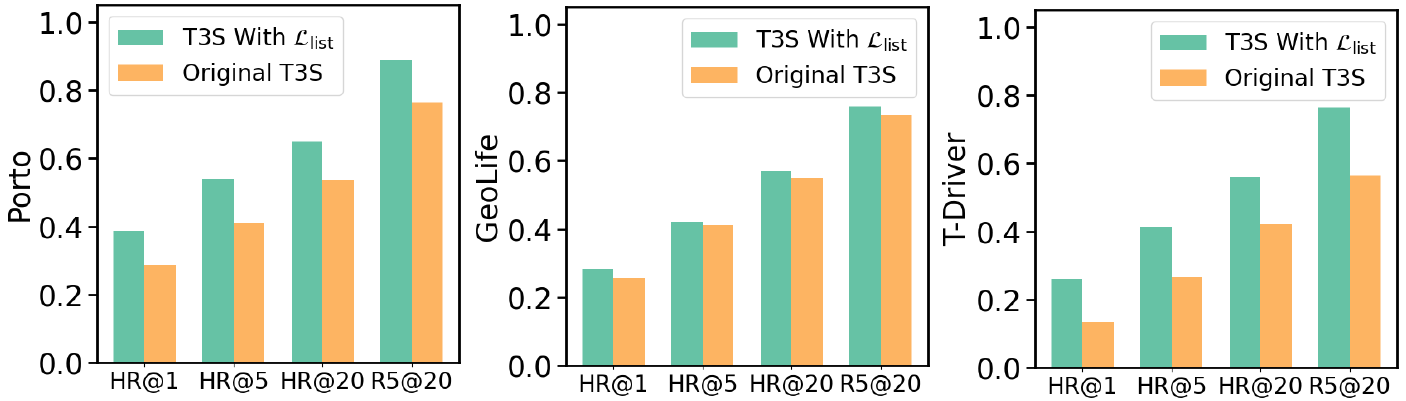} \\
            (a) T3S \\
        \end{tabular}
    \end{minipage}
    \begin{minipage}[t]{1.0\linewidth}
    \centering
        \begin{tabular}{@{\extracolsep{\fill}}c@{}@{\extracolsep{\fill}}}
            \includegraphics[width=1\linewidth]{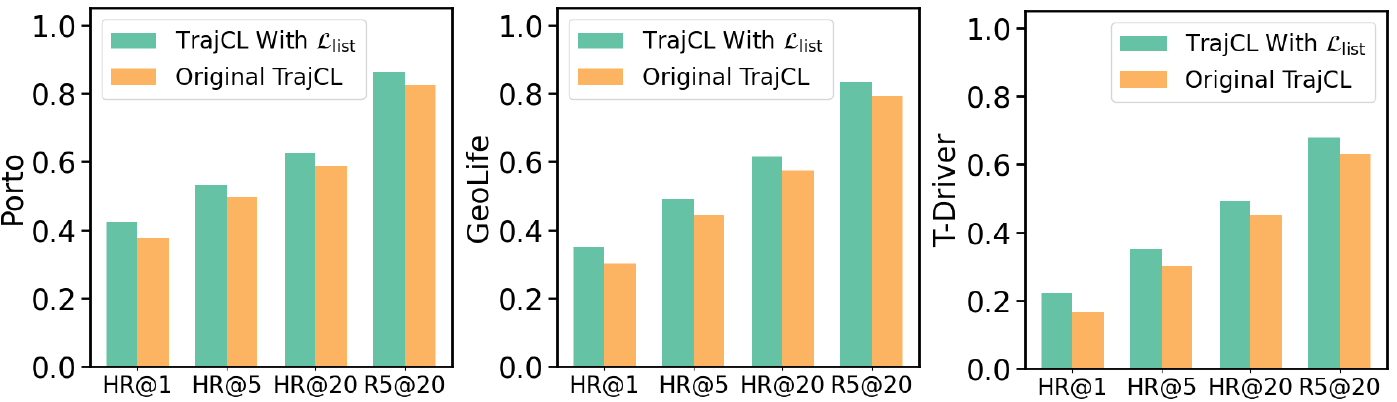}\\
            (b) TrajCL\\
        \end{tabular}
    \end{minipage}
    \caption{Performance improvement T3S/TrajCL with our $\mathcal{L}_{\mathrm{list}}$.}
    \label{fig:pluslistres}
\end{figure}

\begin{figure}[htbp]
    \centering
    \includegraphics[width=\linewidth]{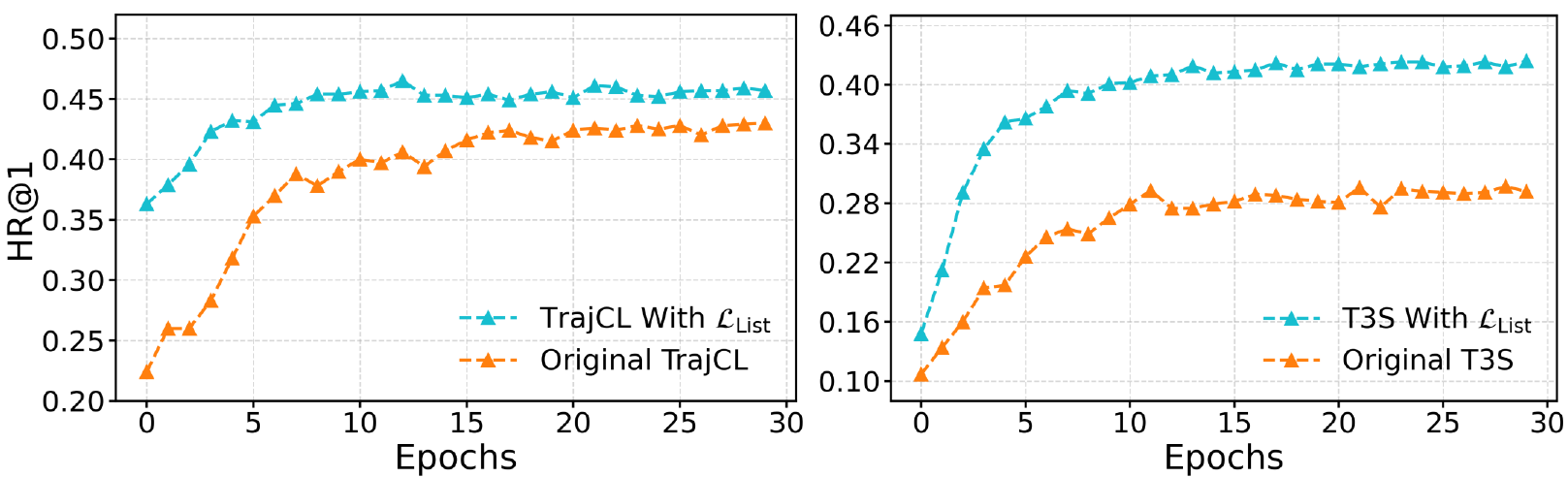}
    \caption{Convergence speed of T3S/TrajCL with our $\mathcal{L}_\text{List}$.}
    \label{fig:pluslist}
\end{figure}

Finally, most existing methods primarily rely on point-wise or list-wise losses as supervision signals, overlooking the global ranking structure across the entire trajectory dataset. In contrast, our approach explicitly integrates global ranking information via the proposed overall ranking-aware regularization, which contributes significantly to its superior performance over all baseline methods. To further verify the effectiveness of overall ranking-aware regularization, we incorporate it into the training processes of TrajCL and T3S by augmenting their loss functions, and evaluate its performance under the SSPD distance metric. As shown in Fig.~\ref{fig:pluslistres}, the overall ranking-aware regularization leads to consistent improvements in the HR@1 across all three datasets, with gains of 23.52\%, 55.69\%, and 21.24\%, respectively. In addition to accuracy improvements, Fig.~\ref{fig:pluslist} also shows that the incorporation of overall ranking-aware regularization accelerates model convergence, reducing the number of training epochs required.

Overall, these findings underscore the general applicability and effectiveness of our proposed method. Furthermore, the consistent improvements observed when integrating our components into existing models provide strong evidence of their standalone utility.


\begin{table}[]
\caption{Comparison of different feature fusion methods. \textit{Abbr.:} AM: Attention Mask.}
\renewcommand{\arraystretch}{1.1}
\label{tab:fusion-method}
\begin{tabular}{c|c|cccc}
\hline
Metric  & Method  & HR@1 & HR@5  & \multicolumn{1}{l}{HR@20} & \multicolumn{1}{l}{R5@20} \\ \hline
\multirow{3}{*}{Hausdorff}   & Sum & 0.359 & 0.518 & 0.641  & 0.859 \\
                                  & AM & 0.414 & 0.553  & 0.647  & 0.898   \\
                                  & ours  & \textbf{0.583} & \textbf{0.685} & \textbf{0.747}  & \textbf{0.969}  \\ \hline
\multirow{3}{*}{Discrete Fr\'echet} & Sum  & 0.370  & 0.515  & 0.654 & 0.843  \\
                                  & AM & 0.409  & 0.558  & 0.705 & 0.902  \\
                                  & ours & \textbf{0.546} & \textbf{0.661} & \textbf{0.759} & \textbf{0.968}  \\ \hline
\multirow{3}{*}{SSPD}             & Sum & 0.388 & 0.539 & 0.649  & 0.887  \\
                                  & AM & 0.375 & 0.502 & 0.602 & 0.838  \\
                                  & ours & \textbf{0.486} & \textbf{0.573} & \textbf{0.658} & \textbf{0.894}  \\ \hline
\end{tabular}
\end{table}

\subsection{Ablation Study (RQ2)} 
To evaluate the effectiveness of the key components, we conduct a detailed ablation experiment on \model by removing different components to get different variants.
\begin{itemize}
\setlength{\itemindent}{-1em}
    \item \textbf{w/o Bridge}: remove the DDBM pre-train.
    \item \textbf{w/o SAM}: remove the semantic alignment module and use the vanilla Transformer instead.
    \item \textbf{w/o $\mathcal{L}_\text{list}$}: remove the overall ranking-aware regularization and use only the MSE loss.
    \item \textbf{w/o $\mathbf{Z}_\text{gps}$}: only return $\mathbf{Z}_{\text{grid}}$ as the trajectory final embedding.
    \item \textbf{w/o $\mathbf{Z}_\text{grid}$}: only return $\mathbf{Z}_{\text{gps}}$ as the trajectory final embedding.
\end{itemize}

\begin{figure}[thbp!]
    \centering
    \begin{minipage}[t]{1.0\linewidth}
    \centering
        \begin{tabular}{@{\extracolsep{\fill}}c@{}@{\extracolsep{\fill}}}
            \includegraphics[width=\linewidth]{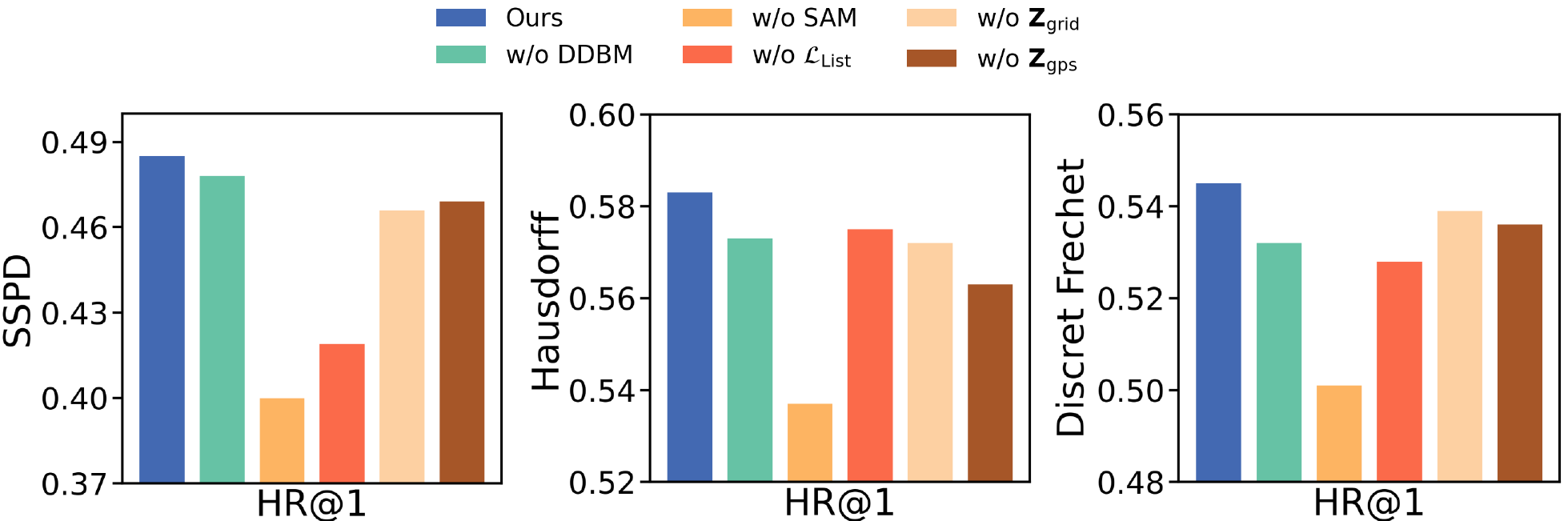} \\
            {\small (a) Porto.} \\
        \end{tabular}
    \end{minipage}
    \begin{minipage}[t]{1.0\linewidth}
    \centering
        \begin{tabular}{@{\extracolsep{\fill}}c@{}@{\extracolsep{\fill}}}
            \includegraphics[width=1\linewidth]{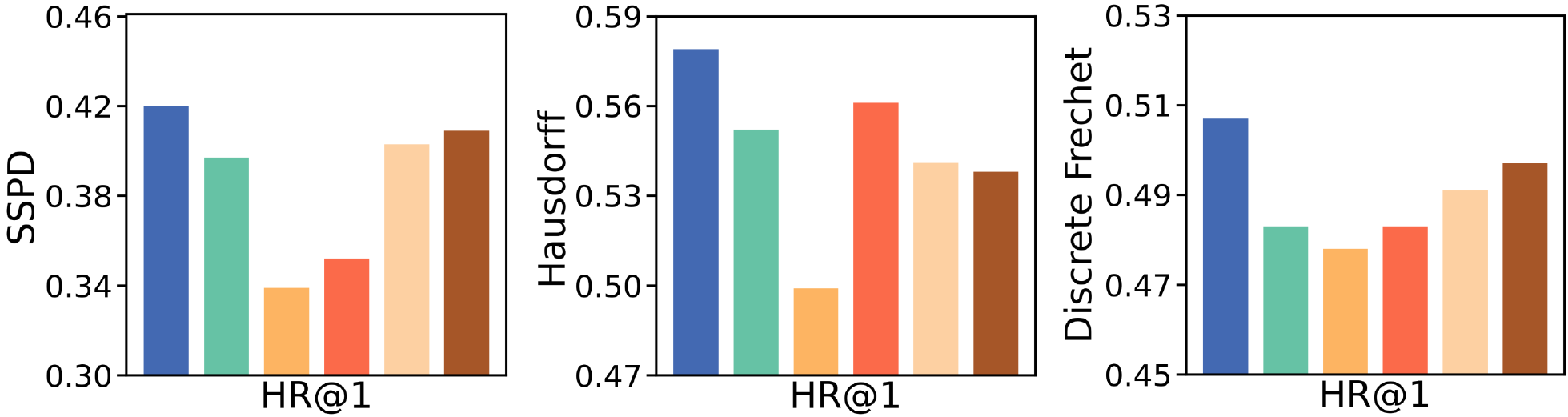}\\
            {\small (b) GeoLife.}\\
        \end{tabular}
    \end{minipage}
    \begin{minipage}[t]{1.0\linewidth}
    \centering
        \begin{tabular}{@{\extracolsep{\fill}}c@{}@{\extracolsep{\fill}}}
            \includegraphics[width=1\linewidth]{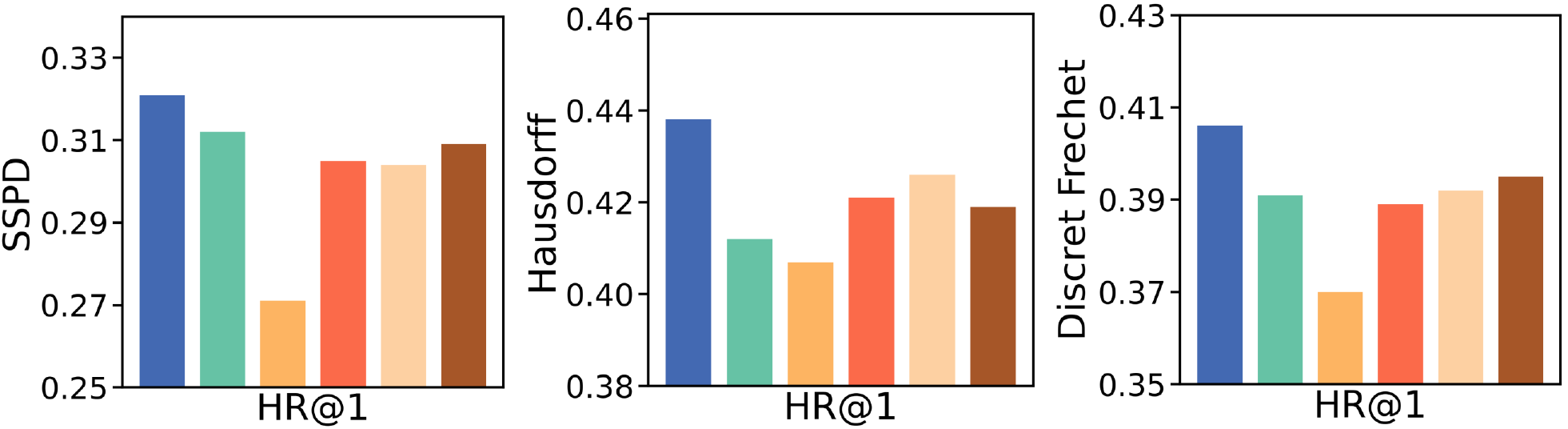}\\
            {\small (c) T-Driver.}\\
        \end{tabular}
    \end{minipage}
    \caption{Ablation study by removing different components.}
    \label{fig:xiaorong}
\end{figure}

The results of the ablation study are shown in Fig.~\ref{fig:xiaorong}. The performance of the model significantly declined after removing each module, demonstrating the effectiveness of each module.

First, \textbf{w/o SAM} shows the most pronounced performance degradation across all heuristic evaluation metrics and datasets. For example, on the Porto dataset, the HR@1 score under the SSPD distance drops by 11.8\%, highlighting the pivotal role of SAM in trajectory similarity learning. This substantial decline demonstrates that removing SAM significantly hampers the model’s ability to align representations across granularities. Notably, naive fusion strategies fail to capture the complex semantic correspondences between coarse-grained and fine-grained views. Without explicitly modeling cross-scale interactions, the model struggles to generate embeddings that faithfully reflect true trajectory similarity.

Second, the performance of \textbf{w/o DDBM} shows a more pronounced decline on the GeoLife and T-Drive datasets, both of which contain higher levels of noise, compared to the Porto dataset. This highlights the crucial role of DDBM-based pretraining in enhancing model robustness under noisy conditions. By simulating a stochastic bridge process between any pair of trajectories, DDBM enables the model to learn noise-resilient pairwise representations. More importantly, our pretraining loss is computed in the high-dimensional embedding space, rather than in the original data space. This design allows the model to bypass local distortions or noise in raw trajectories and focus instead on optimizing meaningful representations in the learned feature space.

In addition, $\textbf{w/o $\mathcal{L}_\text{List}$}$  significantly compromises the effectiveness of the SSPD evaluation metric compared to Hausdorff and Discrete Fréchet distances, resulting in an average decrease of 18.87\% in HR@1 across the three datasets. This is because SSPD does not require precise point-to-point alignment but instead focuses on whether trajectories exhibit similar global path patterns. In comparison, our overall ranking-aware constraint is supervised with ranking as the optimization objective, which is highly aligned with the global trajectory ranking measured by SSPD. Finally, both $\textbf{w/o $\mathbf{Z}_\text{grid}$}$ and $\mathbf{Z}_\text{gps}$ lead to a moderate decline in performance. This is because, compared to the dual semantic alignment attention, a single semantic alignment attention can only achieve limited alignment. It fails to fully explore the bidirectional interaction across different granularities and overlooks the potential impact of reversing their roles.

\subsection{Impact of Hyper-parameters (RQ3)}
\begin{figure}[thbp!]
    \centering
    \begin{minipage}[t]{1.0\linewidth}
    \centering
        \begin{tabular}{@{\extracolsep{\fill}}c@{}@{\extracolsep{\fill}}}
            \includegraphics[width=\linewidth]{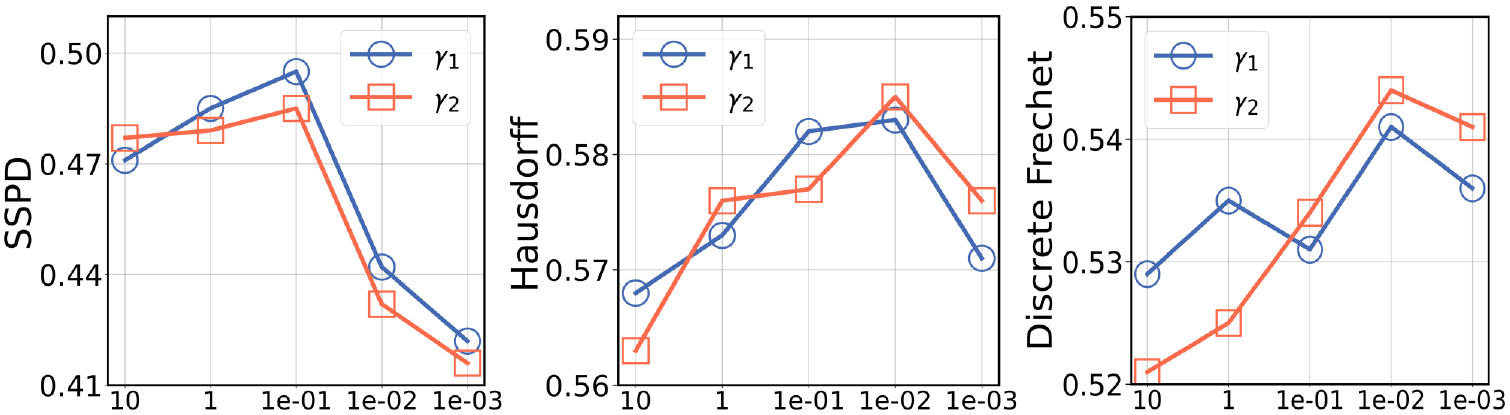} \\
            {\small (a) Fusion weights of $\mathcal{L}_\text{listmle} (\gamma_1)$ and $\mathcal{L}_\text{listnet} (\gamma_2)$.} \\
        \end{tabular}
    \end{minipage}
    \begin{minipage}[t]{1.0\linewidth}
    \centering
        \begin{tabular}{@{\extracolsep{\fill}}c@{}@{\extracolsep{\fill}}}
            \includegraphics[width=1\linewidth]{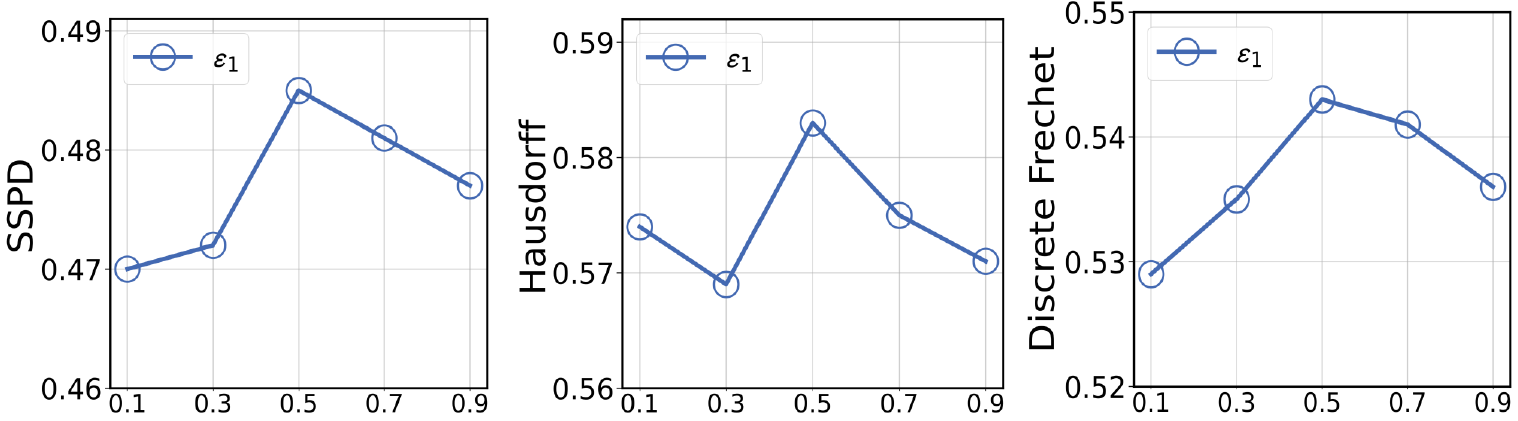}\\
            {\small (b) Fusion weights $\epsilon$.} \\
        \end{tabular}
    \end{minipage}
    \begin{minipage}[t]{1.0\linewidth}
    \centering
        \begin{tabular}{@{\extracolsep{\fill}}c@{}@{\extracolsep{\fill}}}
            \includegraphics[width=1\linewidth]{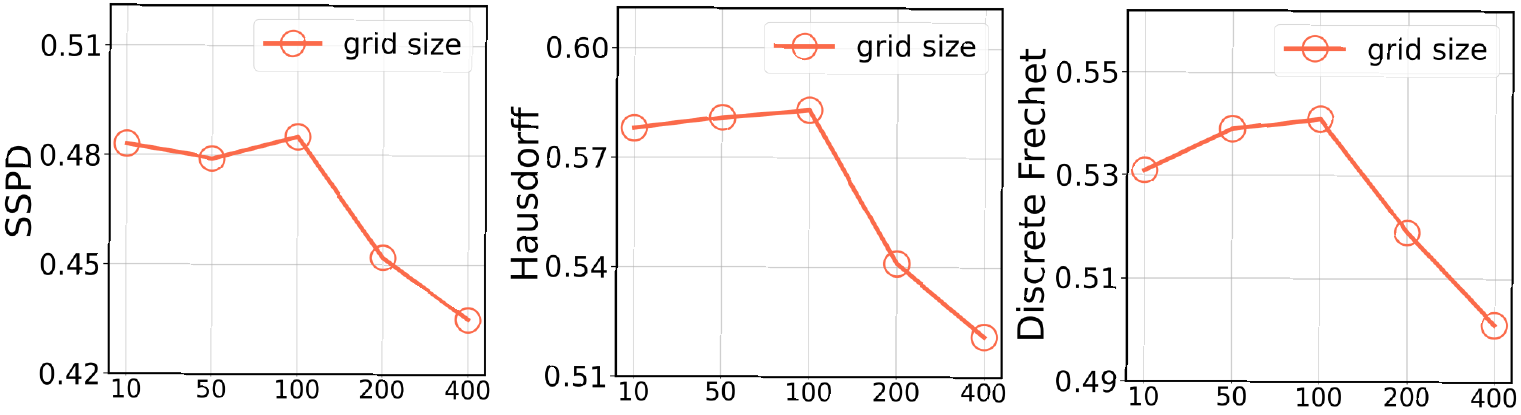}\\
            {\small (c) Grid Size.}\\
        \end{tabular}
    \end{minipage}
    \caption{The impact of hyperparameters on the performance of \model.}
    \label{fig:mingan}
    \vspace{-4mm}
\end{figure}

As shown in Fig.~\ref{fig:mingan}, we further conduct a sensitivity analysis of critical hyperparameters on three evaluation metrics of the Porto dataset and report the results on the HR@1 metric. Initially, larger values of $\gamma_1$ and $\gamma_2$ cause the model to overlook non-global information. As these parameters decrease, a balance between local ranking information and global information is achieved, resulting in improved performance. However, when $\gamma_1$ and $\gamma_2$ become too small, their influence is neglected by the model, leading to a decline in performance.

The performance is optimal when $\mu = 0.5$. This is because the two embeddings represent complementary information, and balancing them maximizes their complementary nature, allowing the model to comprehensively understand different aspects of the data. Over-relying on one aspect of the information may cause the model to lose other important features.

When the initial cell size is small, for example, less than 100, changing the grid size does not significantly affect the model's performance. This is because smaller grids contain less overall trajectory structure information. However, excessively large grids can lead to the loss of structural information. Therefore, the model performs optimally when the cell size is set to 100.

\subsection{Efficiency Comparison (RQ5)}

\subsubsection{Time Complexity Analysis}

To support various distance metrics, our framework employs different pre-encoding strategies: a simple linear projection for SSPD and Hausdorff distances, and an LSTM encoder for the discrete Fréchet distance. For a unified and conservative time complexity analysis, we focus on the LSTM-based pre-encoding, as it represents the most computationally demanding scenario.

The LSTM encoder processes sequences of length $n$ with an embedding dimension $d$, resulting in a time complexity of $\mathcal{O}(n \cdot d^2)$. This reflects the recurrent nature of LSTMs, where each time step involves matrix multiplications of size $d \times d$.

Following pre-encoding, SAM is applied, which comprises several key components: \textbf{(1) Linear projections:} for query, key, and value, each with complexity $\mathcal{O}(n\cdot d^2)$. \textbf{(2) Attention score computations:} including both self- and cross-attention, with complexity $\mathcal{O}(n^2\cdot d)$ due to pairwise interactions between sequence elements. \textbf{(3) Feed-forward networks:} applied to each sequence element, with complexity $\mathcal{O}(n \cdot d \cdot d_{\text{hid}})$, where $d_{\text{hid}}$ denotes the hidden dimension. Aggregating the complexities of all components, the overall time complexity for a single pass through the module is $\mathcal{O}(n \cdot d^2 + n^2 \cdot d + n \cdot d \cdot d_{\text{hid}})$. Here, $n$ is the trajectory length, $d$ is the embedding dimension, and $d_{\text{hid}}$ is the hidden size of the feed-forward layers. This analysis demonstrates that the most computationally intensive part arises from the attention mechanism, especially for longer sequences, while the feed-forward and projection layers contribute linearly for $n$.

\subsubsection{Inference Speed Comparison}

\begin{figure}[htbp]
    \centering
    \includegraphics[width=0.85\linewidth]{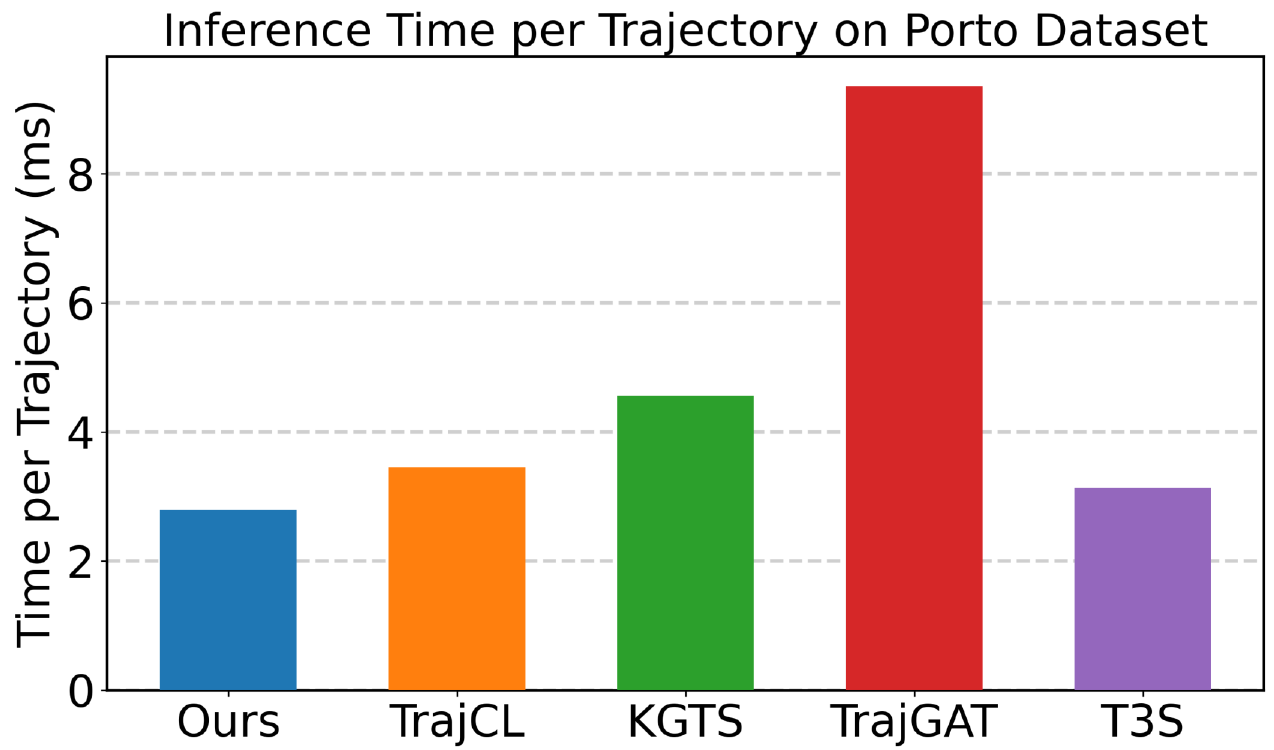}
    \caption{Comparison of inference speed across diverse models.}
    \label{fig:tuilispeed}
\end{figure}

We compare the inference time of our method with several baseline approaches on the Porto dataset, as shown in Fig.~\ref{fig:tuilispeed}. Our model achieves a speed improvement of 10.86\% compared to the second-fastest method, demonstrating its superior efficiency. Compared to TrajCL, which employs two separate Transformer encoders to process spatial and structural information independently and sets the model depth to 2, our model is significantly faster. This efficiency gain stems from our unified architecture that avoids redundant computations across modalities, and from the use of a single-layer SAM, which further reduces computational overhead without sacrificing performance. Although T3S shares a similar architecture with our model, its LSTM input dimension is equal to the hidden dimension of our LSTM. This leads to increased computational overhead for T3S and contributes to its slower inference speed. KGTS and TrajGAT face significant computational challenges due to the large geographic area involved. As the geographical region expands, it leads to the creation of numerous sub-grids. This results in the graph neural network computations involving a vast number of edges and nodes, which substantially increases the time consumption.
These results confirm that our model strikes a favorable balance between representational capacity and computational efficiency, making it well-suited for real-world deployment scenarios where both accuracy and speed are critical.

\begin{figure}[htbp]
    \centering
    \includegraphics[width=\linewidth]{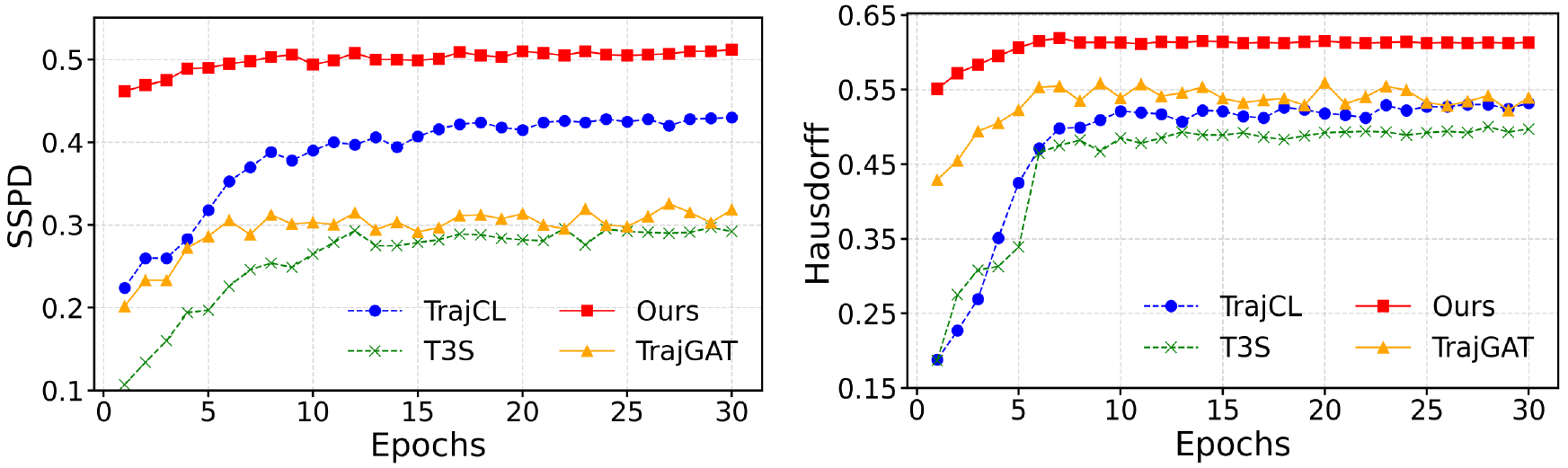}
    \caption{Convergence speed of different methods.}
    \label{fig:converge}
\end{figure}

\subsubsection{Convergence Speed Comparison}
We evaluate and compare the convergence speeds of various representative trajectory learning methods on the Porto dataset, utilizing three widely adopted evaluation metrics to ensure a comprehensive and fair assessment, as shown in Fig.~\ref{fig:converge}. The experimental results clearly demonstrate that our proposed method not only converges significantly faster but also achieves a consistently higher final performance when compared to state-of-the-art baselines, including TrajCL, T3S, and TrajGAT. In particular, our method exhibits a rapid performance gain within the initial few training epochs, reaching a stable state with minimal fluctuations and superior metric scores. This contrasts sharply with the baseline methods, which tend to require a substantially larger number of epochs to achieve convergence. The accelerated convergence observed in our method can be primarily attributed to the effectiveness of our pretraining strategy, which enables the model to capture semantically rich and structurally meaningful trajectory representations from the outset. By initializing the model with well-informed parameters, our approach reduces the burden of learning from scratch and thereby shortens the training cycle.

\section{Conclusion}
In this paper, we propose a novel trajectory similarity computation framework, \model. The model incorporates a semantic alignment module to bridge the semantic gap between features at two distinct scales, integrating them into a unified and robust embedding. Additionally, a denoising diffusion bridge model is employed to pretrain the model in the embedding space, enabling it to learn the interactions and transfer patterns between arbitrary pairs of trajectories, thereby enhancing its robustness and generalization capability. Furthermore, we introduce a global ranking-aware constraint to improve performance on ranking-sensitive evaluation metrics. Experimental results on three public datasets demonstrate the effectiveness and superiority of our method in trajectory similarity computation.



\newpage
\clearpage

\normalem
\bibliographystyle{IEEEtran}
\bibliography{IEEEabrv,sample}

\begin{thebibliography}{10}
\providecommand{\url}[1]{#1}
\csname url@samestyle\endcsname
\providecommand{\newblock}{\relax}
\providecommand{\bibinfo}[2]{#2}
\providecommand{\BIBentrySTDinterwordspacing}{\spaceskip=0pt\relax}
\providecommand{\BIBentryALTinterwordstretchfactor}{4}
\providecommand{\BIBentryALTinterwordspacing}{\spaceskip=\fontdimen2\font plus
\BIBentryALTinterwordstretchfactor\fontdimen3\font minus \fontdimen4\font\relax}
\providecommand{\BIBforeignlanguage}[2]{{%
\expandafter\ifx\csname l@#1\endcsname\relax
\typeout{** WARNING: IEEEtran.bst: No hyphenation pattern has been}%
\typeout{** loaded for the language `#1'. Using the pattern for}%
\typeout{** the default language instead.}%
\else
\language=\csname l@#1\endcsname
\fi
#2}}
\providecommand{\BIBdecl}{\relax}
\BIBdecl

\bibitem{cluster1}
P.~K. Agarwal, K.~Fox, K.~Munagala, A.~Nath, J.~Pan, and E.~Taylor, ``Subtrajectory clustering: Models and algorithms,'' in \emph{Proceedings of the 37th ACM SIGMOD-SIGACT-SIGAI symposium on principles of database systems}, 2018, pp. 75--87.

\bibitem{cluster2}
S.~Wang, Z.~Bao, J.~S. Culpepper, T.~Sellis, and X.~Qin, ``Fast large-scale trajectory clustering,'' \emph{Proceedings of the VLDB Endowment}, vol.~13, no.~1, pp. 29--42, 2019.

\bibitem{traj_cluster}
C.-C. Hung, W.-C. Peng, and W.-C. Lee, ``Clustering and aggregating clues of trajectories for mining trajectory patterns and routes,'' \emph{The VLDB Journal}, vol.~24, pp. 169--192, 2015.

\bibitem{survey}
H.~Su, S.~Liu, B.~Zheng, X.~Zhou, and K.~Zheng, ``A survey of trajectory distance measures and performance evaluation,'' \emph{The VLDB Journal}, vol.~29, pp. 3--32, 2020.

\bibitem{intelligent-transportation1}
D.~Wang, J.~Zhang, W.~Cao, J.~Li, and Y.~Zheng, ``When will you arrive? estimating travel time based on deep neural networks,'' in \emph{Proceedings of the AAAI conference on artificial intelligence}, vol.~32, no.~1, 2018.

\bibitem{ITS}
J.~Kim and H.~S. Mahmassani, ``Spatial and temporal characterization of travel patterns in a traffic network using vehicle trajectories,'' \emph{Transportation Research Procedia}, vol.~9, pp. 164--184, 2015.

\bibitem{urban-planning}
J.~Yuan, Y.~Zheng, and X.~Xie, ``Discovering regions of different functions in a city using human mobility and pois,'' in \emph{Proceedings of the 18th ACM SIGKDD international conference on Knowledge discovery and data mining}, 2012, pp. 186--194.

\bibitem{liugan}
C.~Yang, Z.~Zhang, Z.~Fan, R.~Jiang, Q.~Chen, X.~Song, and R.~Shibasaki, ``Epimob: Interactive visual analytics of citywide human mobility restrictions for epidemic control,'' \emph{IEEE Transactions on Visualization and Computer Graphics}, vol.~29, no.~8, pp. 3586--3601, 2022.

\bibitem{yao2017trajectory}
D.~Yao, C.~Zhang, Z.~Zhu, J.~Huang, and J.~Bi, ``Trajectory clustering via deep representation learning,'' in \emph{2017 international joint conference on neural networks (IJCNN)}.\hskip 1em plus 0.5em minus 0.4em\relax IEEE, 2017, pp. 3880--3887.

\bibitem{yao2018learning}
D.~Yao, C.~Zhang, Z.~Zhu, Q.~Hu, Z.~Wang, J.~Huang, and J.~Bi, ``Learning deep representation for trajectory clustering,'' \emph{Expert Systems}, vol.~35, no.~2, p. e12252, 2018.

\bibitem{survey1}
S.~Wang, Z.~Bao, J.~S. Culpepper, and G.~Cong, ``A survey on trajectory data management, analytics, and learning,'' \emph{ACM Computing Surveys (CSUR)}, vol.~54, no.~2, pp. 1--36, 2021.

\bibitem{tmn}
P.~Yang, H.~Wang, D.~Lian, Y.~Zhang, L.~Qin, and W.~Zhang, ``Tmn: trajectory matching networks for predicting similarity,'' in \emph{2022 IEEE 38th International Conference on Data Engineering (ICDE)}.\hskip 1em plus 0.5em minus 0.4em\relax IEEE, 2022, pp. 1700--1713.

\bibitem{yichang1}
A.~Belhadi, Y.~Djenouri, J.~C.-W. Lin, and A.~Cano, ``Trajectory outlier detection: Algorithms, taxonomies, evaluation, and open challenges,'' \emph{ACM Transactions on Management Information Systems (TMIS)}, vol.~11, no.~3, pp. 1--29, 2020.

\bibitem{yichang2}
F.~Meng, G.~Yuan, S.~Lv, Z.~Wang, and S.~Xia, ``An overview on trajectory outlier detection,'' \emph{Artificial Intelligence Review}, vol.~52, pp. 2437--2456, 2019.

\bibitem{trajcl}
Y.~Chang, J.~Qi, Y.~Liang, and E.~Tanin, ``Contrastive trajectory similarity learning with dual-feature attention,'' in \emph{2023 IEEE 39th International conference on data engineering (ICDE)}.\hskip 1em plus 0.5em minus 0.4em\relax IEEE, 2023, pp. 2933--2945.

\bibitem{grlstm}
S.~Zhou, J.~Li, H.~Wang, S.~Shang, and P.~Han, ``Grlstm: trajectory similarity computation with graph-based residual lstm,'' in \emph{Proceedings of the AAAI Conference on Artificial Intelligence}, vol.~37, no.~4, 2023, pp. 4972--4980.

\bibitem{distcret_frechet}
T.~Eiter and H.~Mannila, ``Computing discrete fr{\'e}chet distance,'' 1994.

\bibitem{survey2}
Y.~Tao, A.~Both, R.~I. Silveira, K.~Buchin, S.~Sijben, R.~S. Purves, P.~Laube, D.~Peng, K.~Toohey, and M.~Duckham, ``A comparative analysis of trajectory similarity measures,'' \emph{GIScience \& Remote Sensing}, vol.~58, no.~5, pp. 643--669, 2021.

\bibitem{survey3}
H.~Yuan and G.~Li, ``A survey of traffic prediction: from spatio-temporal data to intelligent transportation,'' \emph{Data Science and Engineering}, vol.~6, no.~1, pp. 63--85, 2021.

\bibitem{chen2024kgts}
Z.~Chen, D.~Zhang, S.~Feng, K.~Chen, L.~Chen, P.~Han, and S.~Shang, ``Kgts: contrastive trajectory similarity learning over prompt knowledge graph embedding,'' in \emph{Proceedings of the AAAI Conference on Artificial Intelligence}, vol.~38, no.~8, 2024, pp. 8311--8319.

\bibitem{trajgat}
D.~Yao, H.~Hu, L.~Du, G.~Cong, S.~Han, and J.~Bi, ``Trajgat: A graph-based long-term dependency modeling approach for trajectory similarity computation,'' in \emph{Proceedings of the 28th ACM SIGKDD conference on knowledge discovery and data mining}, 2022, pp. 2275--2285.

\bibitem{node2vec}
A.~Grover and J.~Leskovec, ``node2vec: Scalable feature learning for networks,'' in \emph{Proceedings of the 22nd ACM SIGKDD international conference on Knowledge discovery and data mining}, 2016, pp. 855--864.

\bibitem{prtree}
H.~Samet, ``An overview of quadtrees, octrees, and related hierarchical data structures,'' \emph{Theoretical Foundations of Computer Graphics and CAD}, pp. 51--68, 1988.

\bibitem{sun2019rotate}
Z.~Sun, Z.-H. Deng, J.-Y. Nie, and J.~Tang, ``Rotate: Knowledge graph embedding by relational rotation in complex space,'' \emph{arXiv preprint arXiv:1902.10197}, 2019.

\bibitem{zhou2023grlstm}
S.~Zhou, J.~Li, H.~Wang, S.~Shang, and P.~Han, ``Grlstm: trajectory similarity computation with graph-based residual lstm,'' in \emph{Proceedings of the AAAI Conference on Artificial Intelligence}, vol.~37, no.~4, 2023, pp. 4972--4980.

\bibitem{wang2014knowledge}
Z.~Wang, J.~Zhang, J.~Feng, and Z.~Chen, ``Knowledge graph embedding by translating on hyperplanes,'' in \emph{Proceedings of the AAAI conference on artificial intelligence}, vol.~28, no.~1, 2014.

\bibitem{T3S}
P.~Yang, H.~Wang, Y.~Zhang, L.~Qin, W.~Zhang, and X.~Lin, ``T3s: Effective representation learning for trajectory similarity computation,'' in \emph{2021 IEEE 37th International Conference on Data Engineering (ICDE)}.\hskip 1em plus 0.5em minus 0.4em\relax IEEE, 2021, pp. 2183--2188.

\bibitem{lstm}
S.~Hochreiter, ``Long short-term memory,'' \emph{Neural Computation MIT-Press}, 1997.

\bibitem{lstm1}
D.~E. Rumelhart, G.~E. Hinton, and R.~J. Williams, ``Learning representations by back-propagating errors,'' \emph{nature}, vol. 323, no. 6088, pp. 533--536, 1986.

\bibitem{zhang2020trajectory}
H.~Zhang, X.~Zhang, Q.~Jiang, B.~Zheng, Z.~Sun, W.~Sun, and C.~Wang, ``Trajectory similarity learning with auxiliary supervision and optimal matching,'' 2020.

\bibitem{HHL-Traj}
Y.~Cao, L.~Li, X.~Chen, X.~Xu, Z.~Huang, and Y.~Yu, ``Hypergraph hash learning for efficient trajectory similarity computation,'' in \emph{Proceedings of the 33rd ACM International Conference on Information and Knowledge Management}, 2024, pp. 175--186.

\bibitem{hypergraph1}
Y.~Feng, H.~You, Z.~Zhang, R.~Ji, and Y.~Gao, ``Hypergraph neural networks,'' in \emph{Proceedings of the AAAI conference on artificial intelligence}, vol.~33, no.~01, 2019, pp. 3558--3565.

\bibitem{byol}
J.-B. Grill, F.~Strub, F.~Altch{\'e}, C.~Tallec, P.~Richemond, E.~Buchatskaya, C.~Doersch, B.~Avila~Pires, Z.~Guo, M.~Gheshlaghi~Azar \emph{et~al.}, ``Bootstrap your own latent-a new approach to self-supervised learning,'' \emph{Advances in neural information processing systems}, vol.~33, pp. 21\,271--21\,284, 2020.

\bibitem{ag}
A.~Van Den~Oord, N.~Kalchbrenner, and K.~Kavukcuoglu, ``Pixel recurrent neural networks,'' in \emph{International conference on machine learning}.\hskip 1em plus 0.5em minus 0.4em\relax PMLR, 2016, pp. 1747--1756.

\bibitem{t2vec}
X.~Li, K.~Zhao, G.~Cong, C.~S. Jensen, and W.~Wei, ``Deep representation learning for trajectory similarity computation,'' in \emph{2018 IEEE 34th international conference on data engineering (ICDE)}.\hskip 1em plus 0.5em minus 0.4em\relax IEEE, 2018, pp. 617--628.

\bibitem{seq2seq}
I.~Sutskever, ``Sequence to sequence learning with neural networks,'' \emph{arXiv preprint arXiv:1409.3215}, 2014.

\bibitem{seq2seq1}
K.~Cho, B.~Van~Merri{\"e}nboer, C.~Gulcehre, D.~Bahdanau, F.~Bougares, H.~Schwenk, and Y.~Bengio, ``Learning phrase representations using rnn encoder-decoder for statistical machine translation,'' \emph{arXiv preprint arXiv:1406.1078}, 2014.

\bibitem{cl-tsim}
L.~Deng, Y.~Zhao, Z.~Fu, H.~Sun, S.~Liu, and K.~Zheng, ``Efficient trajectory similarity computation with contrastive learning,'' in \emph{Proceedings of the 31st ACM International Conference on Information \& Knowledge Management}, 2022, pp. 365--374.

\bibitem{simclr}
T.~Chen, S.~Kornblith, M.~Norouzi, and G.~Hinton, ``A simple framework for contrastive learning of visual representations,'' in \emph{International conference on machine learning}.\hskip 1em plus 0.5em minus 0.4em\relax PMLR, 2020, pp. 1597--1607.

\bibitem{moco}
K.~He, H.~Fan, Y.~Wu, S.~Xie, and R.~Girshick, ``Momentum contrast for unsupervised visual representation learning,'' in \emph{Proceedings of the IEEE/CVF conference on computer vision and pattern recognition}, 2020, pp. 9729--9738.

\bibitem{ddpm}
J.~Ho, A.~Jain, and P.~Abbeel, ``Denoising diffusion probabilistic models,'' \emph{Advances in neural information processing systems}, vol.~33, pp. 6840--6851, 2020.

\bibitem{ODE}
Y.~Song, J.~Sohl{-}Dickstein, D.~P. Kingma, A.~Kumar, S.~Ermon, and B.~Poole, ``Score-based generative modeling through stochastic differential equations,'' in \emph{9th International Conference on Learning Representations, {ICLR} 2021}.

\bibitem{doob1}
J.~L. Doob and J.~Doob, \emph{Classical potential theory and its probabilistic counterpart}.\hskip 1em plus 0.5em minus 0.4em\relax Springer, 1984, vol. 262.

\bibitem{doob2}
L.~C.~G. Rogers and D.~Williams, \emph{Diffusions, Markov processes, and martingales: It{\^o} calculus}.\hskip 1em plus 0.5em minus 0.4em\relax Cambridge university press, 2000, vol.~2.

\bibitem{karras2022elucidating}
T.~Karras, M.~Aittala, T.~Aila, and S.~Laine, ``Elucidating the design space of diffusion-based generative models,'' \emph{Advances in neural information processing systems}, vol.~35, pp. 26\,565--26\,577, 2022.

\bibitem{ddbm}
L.~Zhou, A.~Lou, S.~Khanna, and S.~Ermon, ``Denoising diffusion bridge models,'' \emph{arXiv preprint arXiv:2309.16948}, 2023.

\bibitem{listnet}
Z.~Cao, T.~Qin, T.-Y. Liu, M.-F. Tsai, and H.~Li, ``Learning to rank: from pairwise approach to listwise approach,'' in \emph{Proceedings of the 24th international conference on Machine learning}, 2007, pp. 129--136.

\bibitem{geolife}
Y.~Zheng, X.~Xie, W.-Y. Ma \emph{et~al.}, ``Geolife: A collaborative social networking service among user, location and trajectory.'' \emph{IEEE Data Eng. Bull.}, vol.~33, no.~2, pp. 32--39, 2010.

\bibitem{tdriver1}
J.~Yuan, Y.~Zheng, C.~Zhang, W.~Xie, X.~Xie, G.~Sun, and Y.~Huang, ``T-drive: driving directions based on taxi trajectories,'' in \emph{Proceedings of the 18th SIGSPATIAL International conference on advances in geographic information systems}, 2010, pp. 99--108.

\bibitem{neutraj}
D.~Yao, G.~Cong, C.~Zhang, and J.~Bi, ``Computing trajectory similarity in linear time: A generic seed-guided neural metric learning approach,'' in \emph{2019 IEEE 35th international conference on data engineering (ICDE)}.\hskip 1em plus 0.5em minus 0.4em\relax IEEE, 2019, pp. 1358--1369.

\end{thebibliography}


\end{document}